\documentclass[letterpaper, 10 pt, conference]{ieeeconf}  % Comment this line out if you need a4paper

\IEEEoverridecommandlockouts                              % This command is only needed if 
\usepackage{graphics} % for pdf, bitmapped graphics files
\usepackage{epsfig} % for postscript graphics files
\usepackage{times} % assumes new font selection scheme installed
\usepackage{amsmath} % assumes amsmath package installed
\usepackage{amssymb}  % assumes amsmath package installed
\usepackage{caption}
\usepackage{multicol}
\usepackage{multirow}
\title{\LARGE \bf
DeepMLE: A Robust Deep Maximum Likelihood Estimator for Two-view Structure from Motion
}

\author{Yuxi Xiao, Li Li, Xiaodi Li and Jian Yao$^{*}$  %<-this % stops a space
\thanks{*Corresponding Author.}% <-this % stops a space
\thanks{The authors are with the School of School of Remote Sensing and Information Engineering, Wuhan University, Wuhan, Hubei, P.R. China.}
}

\begin{document}

\maketitle
\thispagestyle{empty}
\pagestyle{empty}
\linespread{0.95}

%%%%%%%%%%%%%%%%%%%%%%%%%%%%%%%%%%%%%%%%%%%%%%%%%%%%%%%%%%%%%%%%%%%%%%%%%%%%%%%%
\begin{abstract}

Two-view structure from motion (SfM) is the cornerstone of 3D reconstruction and visual SLAM (vSLAM). Many existing end-to-end learning-based methods usually formulate it as a brute regression problem. However, the inadequate utilization of traditional geometry model makes the model not robust in unseen environments. To improve the generalization capability and robustness of end-to-end two-view SfM network, we formulate the two-view SfM problem as a maximum likelihood estimation (MLE) and solve it with the proposed framework, denoted as DeepMLE. First, we propose to take the deep multi-scale correlation maps to depict the visual similarities of 2D image matches decided by ego-motion. In addition, in order to increase the robustness of our framework, we formulate the likelihood function of the correlations of 2D image matches as a Gaussian and Uniform mixture distribution which takes the uncertainty caused by illumination changes, image noise and moving objects into account. Meanwhile, an uncertainty prediction module is presented to predict the pixel-wise distribution parameters. Finally, we iteratively refine the depth and relative camera pose using the gradient-like information to maximize the likelihood function of the correlations. Extensive experimental results on several datasets prove that our method significantly outperforms the state-of-the-art end-to-end two-view SfM approaches in accuracy and generalization capability.

%Although the neural network can fit most of complex mapping functions in theory, the network is prone to overfitting if the physical model-based inference is not applied, especially for the task of two-view SfM that attempts to recover the original information from limited observations. To improve the generalization capability of end-to-end two-view SfM network, we propose to integrate maximum likelihood estimation (MLE) into the network named  DeepMLE. In our network, instead of using the hand-crafted descriptors or photometric residuals, we propose to take the deep multi-scale correlation maps as the observations of 2D matches decided by ego-motion so as to better depict the correspondence of pixels. In addition, in order to increase the robustness of our framework, we novelly design an uncertainty prediction module to predict the pixel-wise distribution parameters for the correlation which is meant to model the uncertainty caused by illumination changes, image noise and moving objects. Finally, we iteratively refine the depth and relative pose using the gradient-like information to maximize the likelihood function of the observations. Extensive experimental results on several datasets prove that our method significantly outperforms the state-of-the-art end-to-end two-view SfM approaches in accuracy and generalization capability.
%

\end{abstract}

%%%%%%%%%%%%%%%%%%%%%%%%%%%%%%%%%%%%%%%%%%%%%%%%%%%%%%%%%%%%%%%%%%%%%%%%%%%%%%%%
\section{INTRODUCTION}

Two-view structure from motion (SfM) aims to recover the 3D structure (depth map) and motion (relative camera pose) from two consecutive images. As the cornerstone of 3D reconstruction and visual simultaneous localization and mapping (vSLAM), this fundamental techniques is widely applied in high-level tasks like autonomous driving~\cite{autodriving}, augmented/virtual reality~\cite{VR}, and robotics~\cite{robotics,slamcomparison}. 
% which is a cornerstone of both 3D reconstruction and visual simultaneous localization and mapping.

Traditional SfM methods usually follow the pipeline of features extraction~\cite{SURF}, sparse 2D image matching~\cite{orb}, relative camera pose computation~\cite{epipolar,five-points1,Efive-points} and 3D triangulation~\cite{andrew2001multiple}. The dense depth map is hard to attain in this pipeline due to the sparsity of 2D image matches. Therefore, the direct methods have been proposed~\cite{DSO,LSD}, which simultaneously calculate a semi-dense depth map and relative camera pose by minimizing the photometric residuals between two images. However, whatever the feature-based or the direct methods are hard to remain robust in the lighting variable condition and dynamic environments~\cite{Reinforce_feature}.

% However, the photometric residuals are very sensitive to some challenging conditions such as less texture and varying lights. In addition, it is also difficult to design an optimal and robust hand-crafted descriptors for various scenes. So the accurate depth map and relative pose are hard to be recovered only by traditional methods especially in large scenes with complex geometry and appearance changes over time~\cite{Reinforce_feature,pixloc}.

% Due to the sparsity of 2D matches and multiple stages of this pipeline, the dense depth is hard to attain and time consuming is high. 

 To overcome these problems, many deep learning-based two-view SfM methods have been proposed recently. In general, learning-based methods can be divided into two categories. The first class~\cite{superpoint,superglue,li2021generalizing} focuses on explicitly learning to match, and the calculation for the depth map and the relative camera pose is left to traditional algorithms~\cite{andrew2001multiple}. The second one tends to generate a dense monocular depth map and relative camera pose using an end-to-end network~\cite{LsNet,DeepSFM,geoconstrains,monodepth,BAnet}. Compared to the first class, the latter benefits from the end-to-end training and inference, and it is usually more efficient. However, developing a robust and generalizable, end-to-end two-view SfM framework remains a challenging task~\cite{survey1}. Therefore, in this paper, we focus on the discussion of the end-to-end learning-based two-view SfM methods.

% Many end-to-end two-view SfM methods propose to directly regress the depth map and relative pose from two input images using supervised or self-supervised DNNs~\cite{survey1}. Nevertheless, due to the absence of explicit geometry model, the models tend to overfit training datasets. Recently, some works~\cite{DeepSFM,BAnet} successfully integrate geometry models and bundle adjustment into neural networks, but they can not ensure the robustness in the dynamic scenes and require considerable computational cost. 

% Although the deep network can approximate any mapping function in theory with the use of large-scale labeled data, the trained model tends to overfit if the physical model-based inference is not applied, especially for the task of two-view SfM that attempts to estimate the inverse mapping from 2D color images to the original 3D scene with limited observations, as illustrated in Fig.~\ref{fig:introduction}. Obviously, for this ill-posed problem, it is not reasonable to directly regress the depth map and relative pose using labeled datasets. This is also the main reason that the generalization capability of existing end-to-end two-view SfM methods is poor.

To improve the generalization capability and robustness of end-to-end two-view SfM networks, we propose to integrate maximum likelihood estimation (MLE) into our designed architecture. First, we propose to learn the multi-scale correlation volumes as a fine depiction for the correspondence of pixels. 
Then, we formulate the likelihood function of correlations of 2D image matches as a Gaussian and Uniform mixture distribution, and an uncertainty module is designed to predict the parameters of the distribution for each pixel. In this mixture distribution, tiny visual differences or illumination changes can be approximated by the Gaussian distribution, while the outliers of observations caused by extreme illumination change, occlusions and moving objects can be modeled by the Uniform distribution. Last, we solve this MLE problem by a proposed deep network instead of using traditional methods. The proposed deep network iteratively searches the optimal depth map and relative camera pose by maximizing the likelihood of observations. Thus, we do not need to manually design the regularizers or priors. Our model can implicitly learn these information from labeled data. The main contributions of our work can be summarized as:
\begin{itemize}
\item We propose to integrate the MLE into the DNNs to maximize the likelihood of correlations of 2D image matches, where the likelihood function is modeled as the Gaussian and Uniform mixture distribution. 
\item We design an end-to-end DNN to iteratively search the optimal estimation to maximize the likelihood using the gradient-like information, which solves the proposed MLE problem efficiently.
% This design simulates the traditional optimization methods and makes the model more interpretable and generalizable. 
% Additionally, the robustness of observations is enhanced by the proposed deep multi-scale correlation maps.
%
\item The proposed DeepMLE achieves state-of-the-art performance and significantly outperforms existing end-to-end two-view SfM methods in both accuracy and generalization capability.

\end{itemize}
%------------------------------------------------------------------------
\section{RELATED WORKS}

\subsection{Traditional Two-view SfM} 
%------------------------------------------------------------------------
%
Traditional methods commonly recover a sparse monocular depth map and relative camera pose by a set of 2D image matched points. Thus, the key of traditional methods is 2D images points matching. To this end, a lot of hand-crafted and distinctive descriptors (e.g. SIFT~\cite{sift}, ORB~\cite{orb}, SURF~\cite{SURF}) were proposed to find the 2D image matches. Although these descriptors are enabled to attain accurate 2D image matches for keypoints, it is impractical to calculate dense 2D image matches for the limitations of computational cost and time~\cite{orb,orbslam2}. To reduce the computational cost and generate the semi-dense depth map, DSO~\cite{DSO} and LSD\cite{LSD} employed photometric bundle adjustment to recover the semi-dense depth map and relative camera pose simultaneously. In addition, in order to alleviate the influence of photometric noise, DTAM~\cite{DTAM} proposed to minimize the photometric residuals along with a TV-L1 regularization. However, the photometric residuals may fail in the conditions of illumination changes and less texture.
% In some classic pipelines, such as ORB-SLAM~\cite{orbslam}, they only applied the sparse feature points to recover the camera pose.
%the computational cost of dense matching based on these descriptors is high.

% to find a set of accurate 2D image matched points

%
%------------------------------------------------------------------------
\subsection{Learning-based Two-view SfM}
%------------------------------------------------------------------------
%To overcome the drawbacks of traditional two-view SfM approaches, many learning-based approaches have been proposed.
\label{related_works}
Generally, existing learning-based methods can be divided into two types~\cite{survey1} including $\textbf{Partial Learning methods}$ and $\textbf{Whole Learning methods}$. $\textbf{Partial Learning methods}$ leverage the powerful representation ability of DNNs to explicitly learn to match, and the most of rest calculations are finished by classic geometry methods. $\textbf{Whole Learning methods}$ pursue to develop a robust pipeline that is end-to-end during the training and the inference.

\textbf{Partial Learning methods.} These methods use the DNNs to extract deep features and explicitly learn the correspondences of pixels. Such correspondences can be represented as the explicit matches, featuremetric loss or the optical flow. Recent works~\cite{Wang_2021_CVPR,li2021generalizing,withoutposenet} proposed to apply the deep optical flow network to find the 2D image matches, the relative camera pose can be directly calculated from the matched points using classical algorithms~\cite{five-points1,Efive-points}. After that, an extra depth estimation module is applied in their framework to further regress the depth map. Specially, Wang $\emph{et al.}$~\cite{Wang_2021_CVPR} proposed a scale-invariant depth module to recover the depth map. In their module, the scale of predicted relative camera pose is applied to guide the estimation of the depth map. However, the works above can only estimate the relative camera pose and the depth with two separate stages. Simulating the traditional direct methods, BA-Net~\cite{BAnet} applied a DNN to extract dense and distinctive visual features to define the featuremetric loss, and then simultaneously estimated a monocular depth map and relative camera pose via bundle adjustment.

% Inspired by this, our DeepMLE also utilizes the DNN to extract dense features to depict the correlation of 2D image matches. However, differently, our work formulates the two-view SfM as a MLE instead of  

%Therefore, recent works~\cite{Wang_2021_CVPR,li2021generalizing,withoutposenet} proposed to apply the deep optical flow network to find the 2D matches, the relative pose can be directly calculated from the matched points using classical algorithms~\cite{five-points1,Efive-points,8points}. After that, an extra depth estimation module is applied in their framework to further regress the depth map. Specially, Wang $\emph{et al.}$~\cite{Wang_2021_CVPR} proposed a scale-invariant depth module to recover the depth map. In their module, the scale of predicted relative pose is applied to guide the estimation of the depth map.

% BA-Net\cite{BAnet} firstly applied a DNN to extract dense and distinctive visual features to define the featuremetric loss, and then simultaneously estimated a monocular depth map and relative pose via bundle adjustment. However, the pose can be decoupled from depth and be estimated independently by the 2D image matches.

% Although the non-end-to-end methods have better interpretability, the accumulated error in different stages cannot be avoided and it is fairly difficult to achieve global optimal for each module. In addition, the efficiency of non-end-to-end methods is relatively low.

\textbf{Whole Learning methods.} These methods attempt to find an end-to-end mapping from two-view images to original 3D structure and motion. In general, these methods can be categorized as self-supervised and supervised methods. The self-supervised methods~\cite{monodepth,SCsfmlearner,geoconstrains,geoconstrains} usually apply the photometric residuals to train the DNNs. In general, they simply apply the U-Net-like convolutional network~\cite{Unet} to directly regress the depth map and relative camera pose. However, the photo-consistency assumption, which assumes the photometric residuals of two matched points should be small, may be violated by the illumination changes and inherent image noise. In addition, it is difficult to minimize the photometric residuals if the intensity gradient is low and the texture information is poor. Yang $\emph{et al.}$~\cite{D3vo} estimated the affine brightness transformation for two images and photometric uncertainty for each pixel to alleviate the influence of illumination changes. For supervised methods, the ground truth is applied to train the network. 
LS-Net~\cite{LsNet} applied the LSTM-RNN as an optimizer to minimize the photometric residuals iteratively, and they took the Jacobian matrix computed by Tensorflow as the input to regress the update of the depth map and relative camera pose. But the Jacobian matrix calculation is memory and time consuming. DeMoN~\cite{DeMoN} proposed a coarse-to-fine architecture and alternately refined the optical flow, depth and motion. DeepSFM~\cite{DeepSFM} proposed to iteratively and alternately refine the initial depth map and initial relative camera pose offered by DeMoN~\cite{DeMoN}. Differently, our DeepMLE does not require the initial depth or relative camera pose, and meanwhile, our method is beneficial from MLE-based inference which clearly show the likelihood probabilistic of estimations. 
% However, the generalization capability and interpretability of end-to-end methods are poor because it is difficult to integrate the physical model into the the neural networks.
%-----------------------------------------------------------------------
%\vspace{-0.25em}
\section{METHOD}
\begin{figure*}[!htb]
	\centering
	\vspace{0.5em}
	\includegraphics[width=0.96\textwidth]{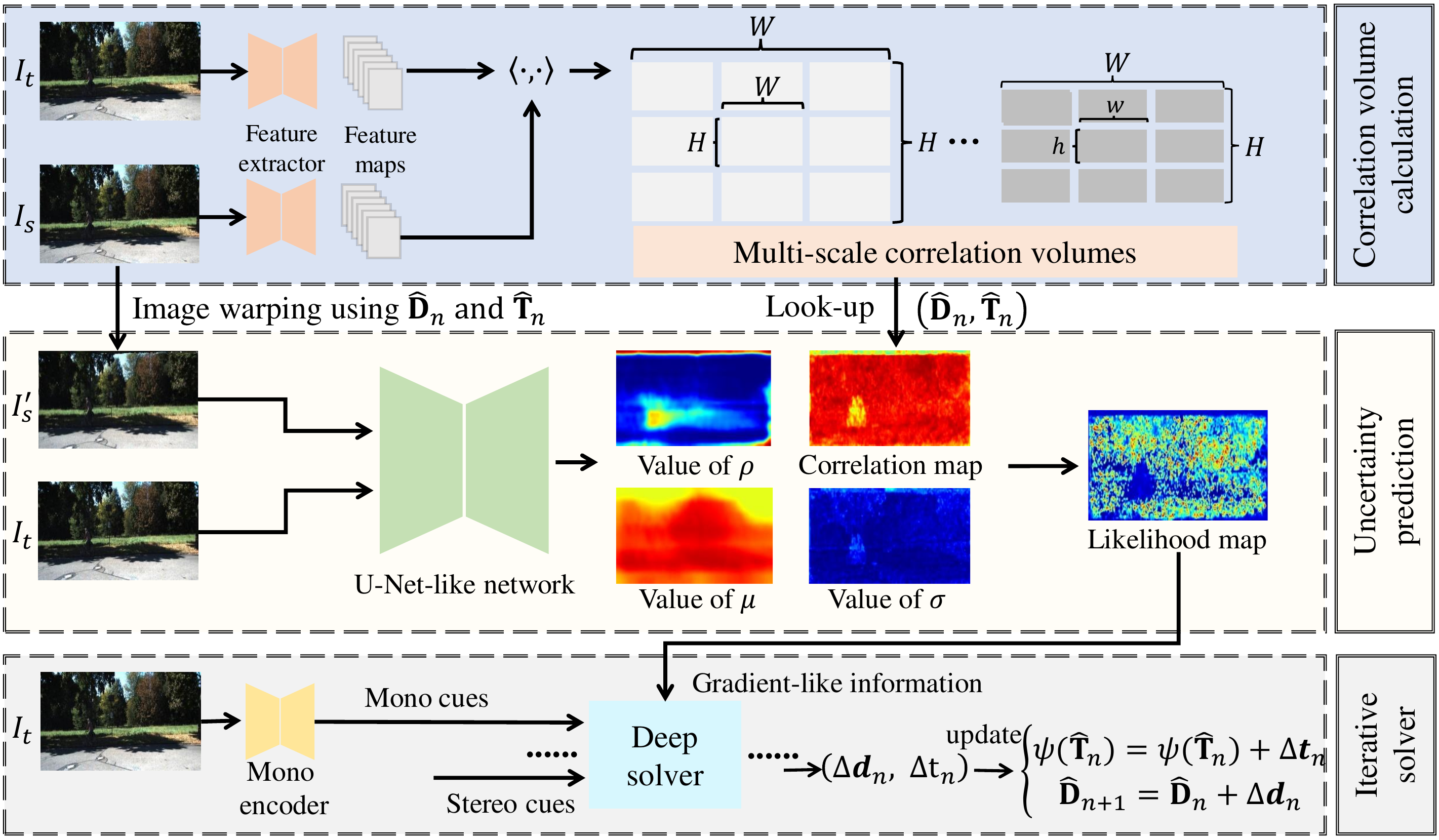}
	\caption{\small The overview of the proposed DeepMLE. At every iteration, the observation of correlations are looked-up in correlation volumes. And then the uncertainty prediction module predicts the uncertainty parameters for every pixel. After attaining the correlation map and uncertainty parameters, the likelihood map and gradient information can be computed. Finally, the update of depth and pose are inferred by deep iterative solver with the input of gradient-like information.}
	\vspace{-0.25em}
\label{Fig:overview}
\vspace{-1.5em}
\end{figure*}

In this section, we are going to introduce how the MLE inference is integrated into the networks. To this end, the general maximum likelihood formula is introduced in Section~\ref{sec:mle} firstly, which demonstrates the process of MLE modelling as a whole. Subsequently, our DeepMLE is introduced to implement the MLE model by three sub-modules in Section~\ref{sec:DeepMLE}. The overview of DeepMLE is shown in Fig.~\ref{Fig:overview}.

%

%------------------------------------------------------------------------
\vspace{-0.5em}
\subsection{Maximum Likelihood Formula for Two-view SfM}
\label{sec:mle}
%------------------------------------------------------------------------
% However, the photometric residuals are sensitive. 
In the two-view SfM problem, given two consecutive images, the target image $I_t$ and the source image $I_s$, the depth map $\textbf{D}$ of $I_t$ and relative camera pose $\textbf{T}$ from $I_t$ to $I_s$ can be estimated by maximizing the likelihood of the correlations. In traditional methods, minimizing the photometric residuals also provide a maximum likelihood solution for the depth and relative camera pose estimation~\cite{DSO,LSD}. In order to increase the robustness of the observations, we take the pixel-wise deep correlations as the observations for estimation $(\widehat{\mathbf{D}},\widehat{\mathbf{T}})$. Meanwhile, we use $C(I_t, I_s)$ (see Section~\ref{sec:corr}) to denote the domain of deep correlation observations, which stores the all-pair correlation values in the target image $I_t$ and the source image $I_s$. For each pixel-wise observation $c$, its value can be looked-up in $C(I_t, I_s)$. The aim of our work is to find the optimal depth map $\mathbf{D}$ and relative camera pose $\mathbf{T}$ that can maximize the likelihood of observations. The definition is formulated as:
% photometric residuals which are the intensity differences of matched pixels are a kind of observations
\begin{equation}
    \{\mathbf{D}, \mathbf{T}\} = \mathop{\arg\max}\limits_{\widehat{\mathbf{D}},\widehat{\mathbf{T}}} \prod\nolimits_{i = 1}^{V}P(\left. c_i \right |\widehat{\mathbf{D}},\widehat{\mathbf{T}}),
\label{eq:likelihood}
\end{equation}
where $V$ is the pixel number of the target image $I_t$. For each pixel in $I_t$, its correlation observation with its correspondence in $I_s$ can be approximated by a Gaussian distribution when only relatively tiny illumination changes happen, but it more likely follow a Uniform distribution if the extreme illumination changes, occlusions and objects moving appear. Therefore, we assume that the observations follow a Gaussian and Uniform mixture distribution. Thus, for each observation $c_i$, the likelihood can be calculated as:
\begin{equation}
    P(\left. c_i \right | \widehat{\mathbf{D}}, \widehat{\mathbf{T}})=(1-\rho_i)\mathcal{N}(\left.c_i\right| \mu_i, \sigma_i)+\rho_i\mathcal{U}(\left.c_i\right|-1,1),
    \label{eq:distribution}
\end{equation}
where $\mu_i$ and $\sigma_i$ denote the mean and standard deviation, respectively, and $\rho_i$ represents the weight between Gaussian and Uniform distributions (see Section~\ref{sec:uncertainty}). 

%------------------------------------------------------------------------
\vspace{-0.6em}
\subsection{Architecture of DeepMLE}
\label{sec:DeepMLE}
%------------------------------------------------------------------------
As shown in Fig.~\ref{Fig:overview}, our DeepMLE implements the MLE model discussed in Section~\ref{sec:mle} by three sub-modules: correlation volume calculation, uncertainty prediction and deep iterative solver.

%------------------------------------------------------------------------
\subsubsection{Correlation Volume Calculation}
\label{sec:corr}
%------------------------------------------------------------------------
As we know, the pixel correspondences between the target image $I_t$ and the source image $I_s$ are the bridge between 2D and 3D vision. However, both hand-crafted descriptors and photometric residuals have several inherent drawbacks which decrease the robustness of the algorithms.  

Thus, to increase the robustness of the observations, we propose to apply a DNN to learn the representations for pixel-wise correlations. We firstly extract the dense deep features of the target image $I_t$ and the source image $I_s$ with a weight-shared extractor $\Phi_{\textrm{corr}}$. And $\mathbf{{F}}_{t}$, $\mathbf{{F}}_{s}\in \mathbb{R}^{H{\times}W{\times}D}$ denote the deep features of $I_t$ and $I_s$ (the height $H$ and width $W$ of deep features are a quarter of those of $I_t$, and the channel number $D$ is set to 256). Then, for every pixel in $I_t$, we dot product its normalized deep features vector with the normalized features of every pixel in $I_s$. By this operation, we can construct a 4D correlation volume $\mathbf{V}\in \mathbb{R}^{H{\times}W{\times}H{\times}W}$ for all pairs in $I_t$ and $I_s$.

In addition, to increase the receptive field, we downsample the last two dimension of $\mathbf{V}$ to build a correlation pyramid $\{\mathbf{V}_k\}_{k=0}^{2}$, where $\mathbf{V}_k$ has dimensions of ${W}\times {H}\times \frac{W}{2^{k}}\times \frac{H}{2^{k}}$.  The correlation pyramid is the domain of observations, and it is denoted as $C(I_t,I_s)$. For each pixel $\mathbf{p} \in I_t$, the corresponding pixel position $\mathbf{p}^\prime$ in $I_s$ can be calculated as:
\begin{equation}
\mathbf{p}^\prime \sim \mathbf{K} \widehat{\mathbf{T}}\widehat{\mathbf{D}}(\mathbf{p})\mathbf{K}^{-1}{\mathbf{p}},
\label{eq:camera}
\end{equation}
where $\mathbf{K}$ is the intrinsic of camera, and $\widehat{\mathbf{D}}(\mathbf{p})$ denote the depth of pixel $\mathbf{p}$, while $\widehat{\mathbf{T}}=[\hat{\mathbf{R}}| \hat{\mathbf{t}}]$ is the transform matrix. $\hat{\mathbf{R}}$ is the relative rotation matrix and $\hat{\mathbf{t}}$ is the relative translation vector. Thus, for each pair $(\mathbf{p}, \mathbf{p}^\prime)$, we can directly look up the correlation value from $C(I_t,I_s)$ according to their positions. We can easily generate the correlation map $I_{\textrm{corr}}\in\mathbb{R}^{H\times W}$ and this correlation map $I_{\textrm{corr}}$ is actually the observation for current estimations $(\widehat{\mathbf{D}}, \widehat{\mathbf{T}})$.

%
% \begin{equation}
% \begin{split}
% \mathbf{{F}}_{t} \in \mathbb{R}^{H{\times}W{\times}D} & = \Phi_{\textrm{corr}}(I_t) , \\ 
% \mathbf{{F}}_{s} \in \mathbb{R}^{H{\times}W{\times}D} & = \Phi_{\textrm{corr}}(I_s) ,
% \end{split}
%   \label{equ2}
% \end{equation}
%
% $\Phi_{\textrm{corr}}$ denotes the network for correlation calculation. 

%------------------------------------------------------------------------
\subsubsection{Uncertainty Prediction}
\label{sec:uncertainty}
%------------------------------------------------------------------------

For each pixel pair $(\mathbf{p}, \mathbf{p}^\prime)$, their correlation is influenced by illumination, occlusion or moving objects and image noise. Thus, we design an uncertainty prediction module to estimate the uncertainty parameters, $\{\rho_i, \mu_i, \sigma_i\}$ in Eq.~(\ref{eq:distribution}), for each pixel. Here, we apply a U-Net-like network to predict the uncertainty parameters with the input of the target image $I_t$ and the warped image $I_s^\prime$ of the source image $I_s$. Let $\Phi_{\textrm{unc}}$ denote the uncertainty network, the uncertainty parameters are estimated as:
\begin{equation}
(I_\rho, I_\mu, I_\sigma) = \Phi_{\textrm{unc}} (I_s^\prime, I_t).
\end{equation}
%
% Because we can predict the uncertainty parameters for each pixel in target image $I_t$, so we can obtain three uncertainty parameter maps $I_\rho$, $I_\mu$, $I_\sigma\in\mathbb{R}^{H\times W}$. 

It should be noted that we do not need the ground truth of uncertainty parameter maps to train the network. This module allows the network to adaptively predict the uncertainty using the input data, which improves the accuracy and robustness of the model. 
% From the visual example presented in Fig.~\ref{Fig:uncertainty}, we observe that there is a high reflecting area in $I_t$. The correlation observations of this area more likely follow the Uniform distribution, and $\Phi_{\textrm{unc}}$ predicts the large $\rho$ for this area as we expect. Meanwhile, the values of $\mu$ and $\sigma$ of this area are small because the Gaussian distribution is almost not adopted in this area. In addition, the values of predicted $\sigma$ also obey the fact that the pixels on the boundary are influenced by foreground and background, and thus their intensities or visual features are sensitive to the change of the view, so the predicted $\sigma$ is usually high in this area.  

% \begin{figure}[t]
% \centering
% \renewcommand{\tabcolsep}{2pt}
% \begin{tabular}{cccc}
% \multicolumn{1}{c}{\includegraphics[width= 0.24 \linewidth]{images/uncertainty/6/org}} & \multicolumn{1}{c}{\includegraphics[width= 0.24 \linewidth]{images/uncertainty/6/rho}} &
% \multicolumn{1}{c}{\includegraphics[width= 0.24 \linewidth]{images/uncertainty/6/mean}} & \multicolumn{1}{c}{\includegraphics[width= 0.24 \linewidth]{images/uncertainty/6/sigam}} \\
% $I_t$ & $I_{\rho}$ & $I_{\mu}$ & $I_{\sigma}$
% \end{tabular} 
% \caption{A visual illustration of uncertainty maps. }
% \label{Fig:uncertainty}
% \end{figure}

After introduction of the uncertainty module, for any estimation $(\widehat{\mathbf{D}}, \widehat{\mathbf{T}})$, the corresponding likelihood map $I_{l}$ can be directly calculated according to Eq.~(\ref{eq:distribution}).
%------------------------------------------------------------------------
\subsubsection{Deep Iterative Solver}
%------------------------------------------------------------------------
For solving this MLE problem in an end-to-end way, we design a deep iterative solver which refines the depth and relative camera pose step by step. Inspired by the traditional gradient descent optimization algorithm, the proposed deep solver uses the gradient-like information of the depth map and relative camera pose to iteratively refine the initial estimation. Let $\Phi_{\textrm{updata}}$ denote the network of deep iterative solver. Because the rotation matrix can not be updated by simple addition and subtraction, we first reparametrize the transform matrix $\widehat{\mathbf{T}}_n$ to six parameters $\psi(\widehat{\mathbf{T}}_n)=\{r_x, r_y, r_z, t_x, t_y, t_z\}$, which are the Lie Algebra for the transform matrix. Let $(\widehat{\mathbf{D}}_n, \psi(\widehat{\mathbf{T}}_n))$ denote the current estimation, we need to predict the update $(\Delta \mathbf{d}_n, \Delta \mathbf{t}_n)$ to generate the next estimation $(\widehat{\mathbf{D}}_{n+1}, \psi(\widehat{\mathbf{T}}_{n+1}))$,
\begin{equation}
\widehat{\mathbf{D}}_{n+1} = \widehat{\mathbf{D}}_n + \Delta \mathbf{d}_n, \psi(\widehat{\mathbf{T}}_{n+1}) = \psi(\widehat{\mathbf{T}}_n) + \Delta \mathbf{t}_n.
\end{equation}

% \textbf{Gradient descent method.} To maximize the objective cost function defined in Eq.~(\ref{eq:likelihood}), a traditional and typical method is the gradient descent. Considering a pixel $\mathbf{p} \in I_t$, the final likelihood (energy cost) $I_l(\mathbf{p})$ of corresponding observation is calculated as:
% %
% \begin{equation}
% I_{l}(\mathbf{p}) = P( I_{\textrm{corr}}(\mathbf{p}), (I_\rho(\mathbf{p}), I_\mu(\mathbf{p}), I_\sigma(\mathbf{p}) ),
% \label{Eq:observation}
% \end{equation}
% %
% where $I_{\textrm{corr}}(\mathbf{p})$ denotes the correlation value of $\mathbf{p}$. $(I_\rho(\mathbf{p}), I_\mu(\mathbf{p}), I_\sigma(\mathbf{p})$ denotes the three uncertainty values of $\mathbf{p}$. $P(\cdot)$ represents the Gaussian and Uniform mixture distribution defined in Eq.~(\ref{eq:distribution}).

% To determine the direction of the update, we need to calculate the Jacobian or Hessian matrix of the depth and relative pose. However, because $I_{\textrm{corr}}$ and $\mathcal{I}_{\textrm{unc}}$ are not continuous and computed by deep networks, it is difficult to take the derivative of the depth map and relative pose. Therefore, we propose computing gradient-like information by difference approximation method below. 

\textbf{Gradient-like information.} In order to mimic gradient descent method, we use difference approximation to obtain the first-order derivation of the depth and relative camera pose. For current estimation $(\widehat{\mathbf{D}}_n, \psi(\widehat{\mathbf{T}}_n))$, we give a set of fixed and tiny disturbance $\pm \Delta \mathbf{G} = (\Delta \widehat{\mathbf{D}}_n, \Delta \psi(\widehat{\mathbf{T}}_n))$ to slightly change the value of parameters. It should be noted that we give the same depth disturbance for all pixels to reduce the computational cost. After the disturbance, we can generate a set of new likelihood maps $\mathcal{J}_{l}$ according to Eq.~(\ref{eq:distribution}). Then, for $J_{l}^{m} \in \mathcal{J}_{l}$, the difference likelihood map $\Delta J_{l}^m =  J_{l}^{m} - I_{l}$. The difference likelihood maps contain the gradient-like information for each parameter in $(\widehat{\mathbf{D}}_n, \psi(\widehat{\mathbf{T}}_n))$. So these difference likelihood maps are leveraged in next in update prediction. 

\begin{figure}[t]
	\centering
	\vspace{0.5em}
	\includegraphics[width=0.48 \textwidth]{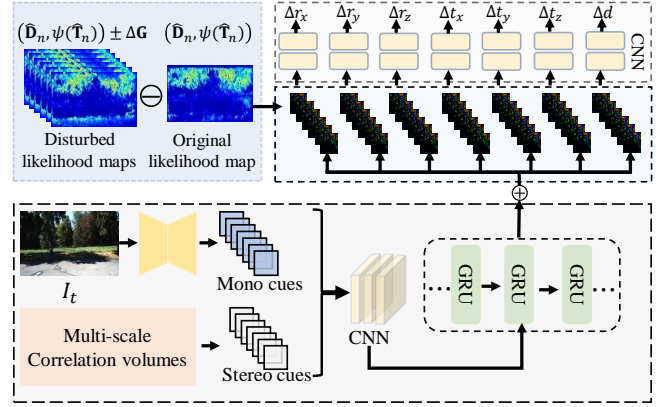}\\[-0.25em]
	\caption{\small The architecture of our proposed deep iterative solver. Disturbed likelihood maps are computed by a set of disturbed depth and relative camera pose. The update of $\textbf{G}$ (depth and relative camera pose) are predicted by seven different CNN heads with the input of difference likelihood map and GRU output.}
	\vspace{-1.5em}
\label{fig:solver}
\end{figure}

\textbf{Update prediction.} We propose to use a gate recurrent unit (GRU)~\cite{GRU} to predict the update $(\Delta \mathbf{d}_n, \Delta \mathbf{t}_n)$. As shown in Fig.~\ref{fig:solver}, the input of GRU includes three parts: the mono cues, the stereo cues and the output of last unit. The mono cues are extracted from the target image $I_t$ using a CNN as the supplementary information for assisting the depth estimation especially in the areas of mutual invisibility. The stereo cues are extracted in the similar way presented in RAFT~\cite{raft}. The output of GRU contains the temporal information of optimization. Then, we concatenate the output feature of GRU with the gradient-like information to predict the optimal gradient descent direction for parameter update using a CNN. After that, we directly obtain the transform matrix $\widehat{\mathbf{T}}_{n+1} = \psi^{-1}(r_x, r_y, r_z, t_x, t_y, t_z)$. By this design, our update prediction module can leverage not only the temporal and also the spatial information of the optimization to give a more accurate prediction. 
% The mono cues are important for recovering the depth map of target image especially for the areas of mutual invisibility. We use the similar way presented in RAFT~\cite{raft} to extract the stereo cues from the correlation volumes. The history information is provided by the latest unit. The difference likelihood maps of the unknown parameters contain the gradient information of different scales. The output of GRU contains the history information of optimization. 
%and the difference likelihood maps as input to predict the update for the depth and the relative camera pose.
% the mono cues, the stereo cues and history information. The mono cues are extracted from the target image $I_t$ using a CNN. The mono cues are important for recovering the depth map of target image especially for the areas of mutual invisibility. We use the similar way presented in RAFT~\cite{raft} to extract the stereo cues from the correlation volumes. The history information is provided by the latest unit. The difference likelihood maps of the unknown parameters contain the gradient information of different scales. The output of GRU contains the history information of optimization. 
%
%

%------------------------------------------------------------------------
\vspace{-0.6em}
\subsection{Loss Function}

%-----------------------------------------------------------------------

Our whole model contains three sub-modules, namely, $\Phi_{\textmd{corr}}, \Phi_{\textmd{unc}}, \Phi_{\textmd{update}}$. Let $\theta_{\textrm{corr}}$, $\theta_{\textrm{unc}}$ and $\theta_{\textrm{update}}$ denote the model parameters of these three sub-modules, respectively. The most widely used loss is the $L_2$ norm ~\cite{DeepSFM,DeepV2D,DeMoN} between prediction $(\widehat{\mathbf{D}}_n, \widehat{\mathbf{T}}_n)$ and ground truth $(\mathbf{D}_{gt},\mathbf{T}_{gt})$, this regression loss $\mathcal{L}_{reg}$ is defined as:
 \begin{equation}\small
    \mathcal{L}_{\textrm{reg}}\!=\!\sum_{n=1}^N(\Vert{\alpha_n}\widehat{\mathbf{D}}_n\!-\!\mathbf{D}_{gt}\Vert_2\!+\!\Vert\widehat{\mathbf{R}}_n\!-\!\mathbf{R}_{gt}\Vert_2\!+\!\Vert{\alpha_n}\hat{\mathbf{t}}_n\!-\!\mathbf{t}_{gt}\Vert_2 ),
\end{equation}
where $N$ is the number of iterations and $\alpha_n$ is the scale factor between ground truth and current estimation. However, just using $\mathcal{L}_{\textrm{reg}}$ can not give adequate and clear constrains for three sub-modules. To give more specific and proper supervision signals for each module, we further propose a probabilistic loss $\mathcal{L}_{\textrm{prob}}$ and a likelihood increase loss $\mathcal{L}_{\textrm{inc}}$.     

$\textbf{Probabilistic loss}$. This loss is designed to find the optimal estimation of ${\theta}_{\textrm{corr}}$ and ${\theta}_{\textrm{unc}}$. We also use the maximum likelihood estimation to estimate ${\theta}_{\textrm{corr}}$ and ${\theta}_{\textrm{unc}}$ as follows:
\begin{equation}
\scriptsize
\begin{split}
& \{{\theta}_{\textrm{corr}},{\theta}_{\textrm{unc}}\} =\mathop{\arg\max}\limits_{\hat{{\theta}}_{\textrm{corr}},\hat{\theta}_{\textrm{unc}}}P(\left.I_{\textrm{corr}}\right|\hat{{\theta}}_{\textrm{corr}},\hat{\theta}_{\textrm{unc}}), \\
& P(\left.c\right|\hat{{\theta}}_{\textrm{corr}}\!,\!\hat{\theta}_\textrm{unc}) \!=\!\! \sum_{n=1}^{N} \!\big(\! \prod_{c \in I_{\textrm{corr}}}\!P( c  |\widehat{\mathbf{D}}_n,\!\widehat{\mathbf{T}}_n,\!\hat{{\theta}}_{\textrm{corr}},\!\hat{\theta}_\textrm{unc})  \!P(\left.\widehat{\mathbf{D}}_n,\!\widehat{\mathbf{T}}_n\right|\textbf{D}_{gt},\!\textbf{T}_{gt}) \!\big)
\end{split}
\end{equation}
We assume that the $P(\left. \widehat{\mathbf{D}}_n,\widehat{\mathbf{T}}_n\right|\textbf{D}_{gt},\textbf{T}_{gt})$ is a impulse-like distribution. The expression above can be converted into a probabilistic loss, 
\begin{equation}
\small
\begin{split}
    \mathcal{L}_{\textrm{prob}}&=-\sum\nolimits_{n=1}^{N}e^{l_n-\frac{\mathcal{L}^n_{\textrm{reg}}}{\sigma}},\\
    l_n&=\frac{\sum_{c \in I_{\textrm{corr}} }log(P(\left. c \right |\widehat{\mathbf{D}}_n,\widehat{\mathbf{T}}_n,\hat{{\theta}}_{\textrm{corr}},\hat{\theta}_{\textrm{unc}})))}{V},
\end{split}
\end{equation}
where $\sigma$ is a hyper-parameter for the impulse-like distribution, and $\mathcal{L}^n_{\textrm{reg}}$ is the regression loss of the $n$-th iteration. $l_n$ denotes the average likelihood of current observations.

$\textbf{Likelihood increase loss}$. In order to give a clear and specific supervision signal to estimate $\theta_{\textrm{update}}$, we design a likelihood increase loss $\mathcal{L}_{\textrm{inc}}$. In this loss, we expect that the likelihood of current iteration can rapidly increase when the different between current estimation and ground truth is large. Otherwise, the likelihood tends to convergence. $\mathcal{L}_{\textrm{inc}}$ is defined as:
\begin{align}
    \mathcal{L}_{\textrm{inc}}&=\sum\nolimits_{n=1}^N(l_n-l_{n+1})log(1+{\mathcal{L}^n_{\textrm{reg}}}).
\end{align} 
Finally, our total loss can be expressed as below,
\begin{equation}
    \mathcal{L}_{\rm total}=\alpha_1\mathcal{L}_{\textrm{reg}}+\alpha_2\mathcal{L}_{\textrm{inc}}+\alpha_3\mathcal{L}_{\textrm{prob}},
    \label{total_loss}
\end{equation}
where $\alpha_1$, $\alpha_2$ and $\alpha_3$ are three hyper-parameters.

%-----------------------------------------------------------------------
\vspace{-0.25em}
\section{EXPERIMENTS}
%------------------------------------------------------------------
In this section, we first introduce the implementation details of our network. Then, we present the evaluation results of our DeepMLE in both accuracy and generalization. Meanwhile, we carefully compare our DeepMLE with the state-of-art $\textbf{Partial Learning methods}$ and $\textbf{Whole Learning methods}$ (Seen in Section\ref{related_works}). And in this section, we denote these two types methods as $\textbf{Type I}$ and $\textbf{Type II}$ methods, respectively. Finally, we give the ablation study to justify our design for each module.
% We evaluated DeepMLE on widely used datasets and compared it with both state-of-the-art $\textbf{Type I}$ and $\textbf{Type II}$ methods. For fair comparison, we scaled our results using the same strategy with \cite{Wang_2021_CVPR}.
% denoted as $\textbf{E2E}$, and non end-to-end methods, denoted as $\textbf{non-E2E}$. For fair comparison, we scaled our results using the same strategy with \cite{Wang_2021_CVPR}. 

% Implementation details (i.e. network setting and training details) are provided in supplementary material.
%------------------------------------------------------------------
\vspace{-0.5em}
\subsection{Implementation Details}
We implemented our framework using PyTorch. For the training on the KITTI dataset~\cite{KITTI_vo}, we trained our model on one NVIDIA RTK 3090 GPU to convergence on all 40k training two-view pairs. In order to accelerate the training, we first initialized our model by training it with low resolution images ($188\times620$). And in initialization, we set the $\alpha_1$, $\alpha_2$, and $\alpha_3$ in the total loss as shown in Eq.~(\ref{total_loss}) to $0.05$, $1$, $0.05$, respectively. After initialization training, we refined our model with high resolution images ($256\times832$). And $\alpha_1$, $\alpha_2$, and $\alpha_3$ was set to $1$, $1$, and $1$, respectively. The number of iterations was set to 8. The total time of training costs 10 days for six epochs. The initialization training was done with a batch size of 2 and a learning rate of $5\times 10^{-4}$ in first two epochs. And the last four epochs are in the refining procedure with a batch size of 1 and a learning rate of $8\times 10^{-5}$. The training optimizer is AdamW~\cite{Adamw}, and the gradients were clipped into $[-1,1]$. For the training in DeMoN~\cite{DeMoN}, the learning rate was set to $5\times 10^{-5}$ in the first eight epochs and the $\alpha_1$, $\alpha_2$, and $\alpha_3$ were set to $0.5$, $2$, and $1$, respectively, and in the last four epochs the learning rate was set to $5\times 10^{-6}$. The number of iterations was also set to 8.
%------------------------------------------------------------------
\vspace{-0.5em}
\subsection{Datasets}
%------------------------------------------------------------------

$\textbf{KITTI.}$ KITTI is a widely used benchmark dataset in automatic driving tasks. For depth evaluation, we conducted the experiments on Eigen split~\cite{eigen_split}. In order to adapt it for evaluating the depth estimation for SfM, we additionally selected the near image of the target image as the source image, which is the same scheme applied in~\cite{Wang_2021_CVPR}. For pose evaluation, we followed the most popular evaluation fashion in end-to-end SfM methods~\cite{geoconstrains,monodepth,Wang_2021_CVPR} to train our model in the first 8 sequences in KITTI odometry and test it in the ``09'' and ``10'' sequences.

$\textbf{vKITTI.}$ vKITTI~\cite{vkitti} contains several sequences of virtual data including different imaging and weather conditions. In order to evaluate the generalization capability of our network, we follow the setting in the domain-adaptation task\cite{S2R_2021_CVPR}. We trained DeepMLE in vKITTI and evaluated the depth estimation in KITTI without any fine-tune. 

$\textbf{DeMoN dataset.}$ DeMoN dataset~\cite{DeMoN} selects the data from various sources including SUN3D~\cite{SUN3D}, RGB-D~\cite{RGBD} and Scenes11~\cite{Scenes11}. It contains both indoor and outdoor scenes. Therefore, this comprehensive dataset is representative in evaluating the performance of the SfM pipeline. And for a fair comparison, we evaluated our methods with DeMoN metrics. Meanwhile, for further illusion of generalization capability of our work, we also tested DeepMLE on MVS dataset~\cite{MVS} which do not appear in our training data.

\begin{figure*}[!htb]
\centering
\vspace{0.5em}
\small
\renewcommand{\tabcolsep}{2pt}
\begin{tabular}{cccc}
\multicolumn{1}{c}{\includegraphics[width= 0.24 \linewidth]{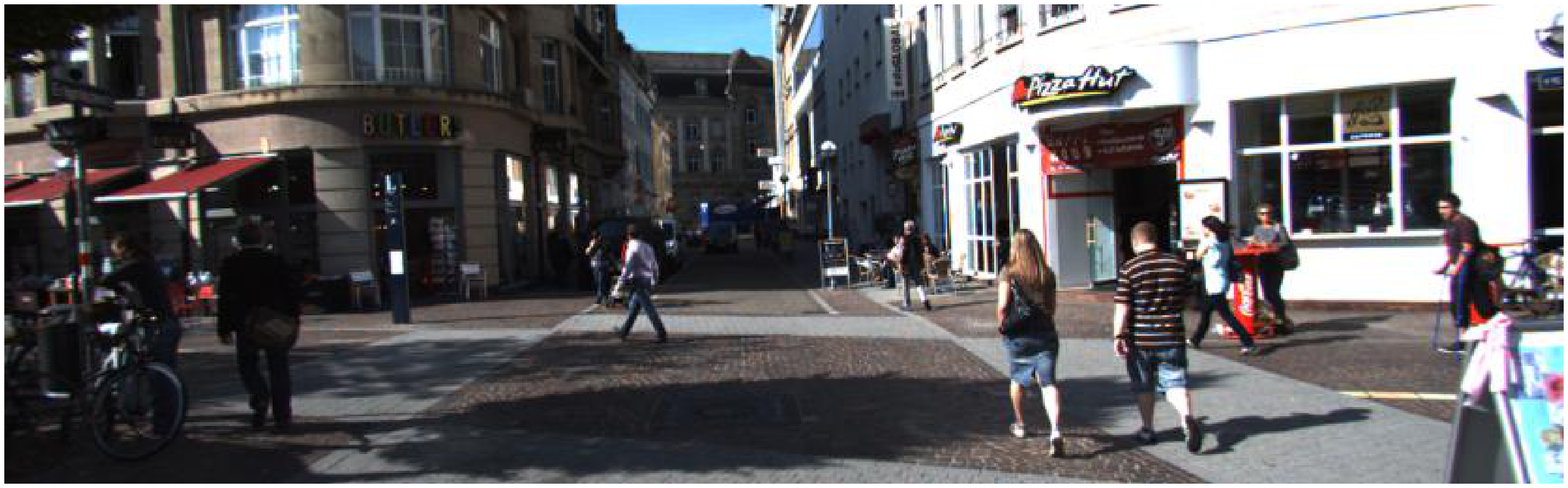}} & \multicolumn{1}{c}{\includegraphics[width= 0.24 \linewidth]{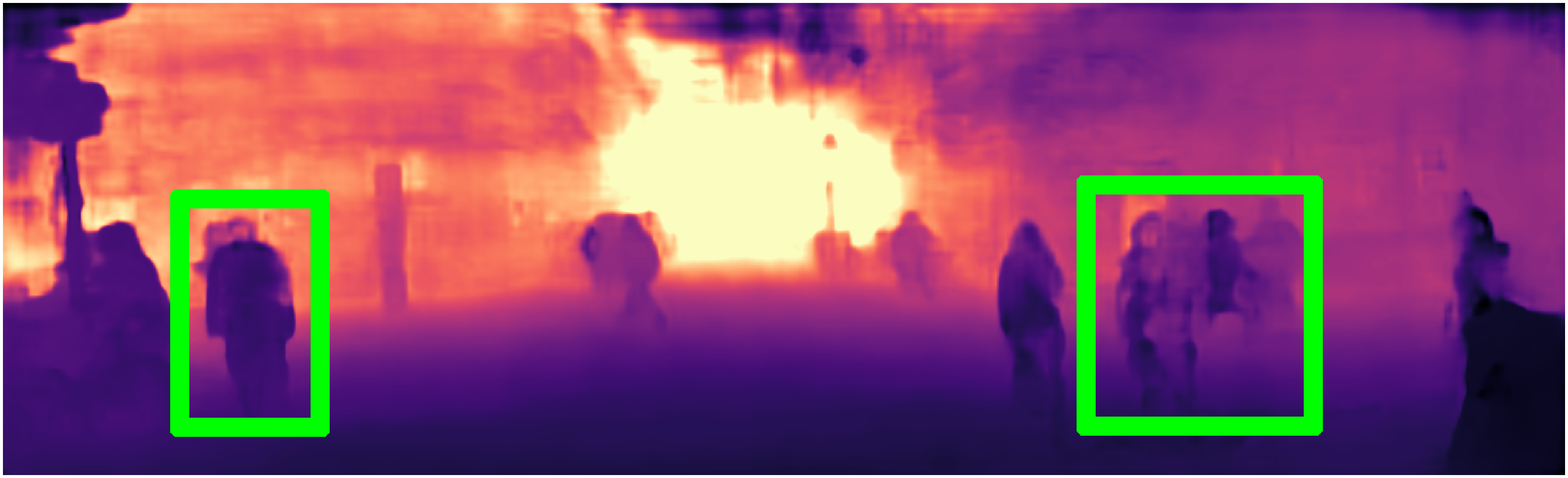}} &
\multicolumn{1}{c}{\includegraphics[width= 0.24 \linewidth]{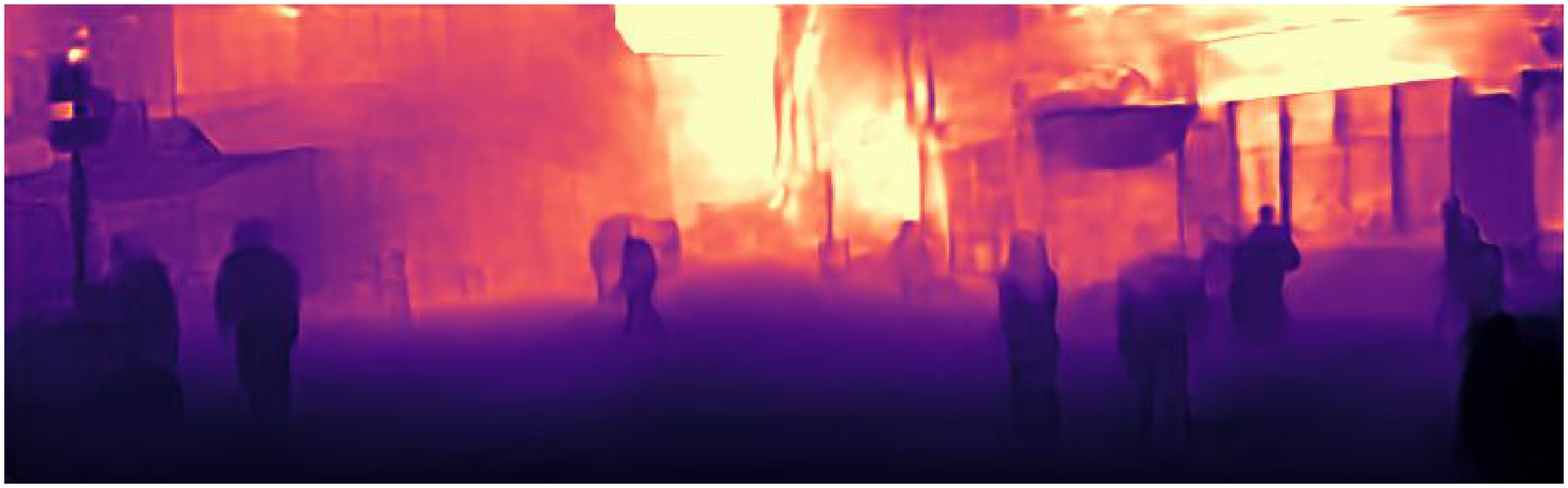}} & \multicolumn{1}{c}{\includegraphics[width= 0.24 \linewidth]{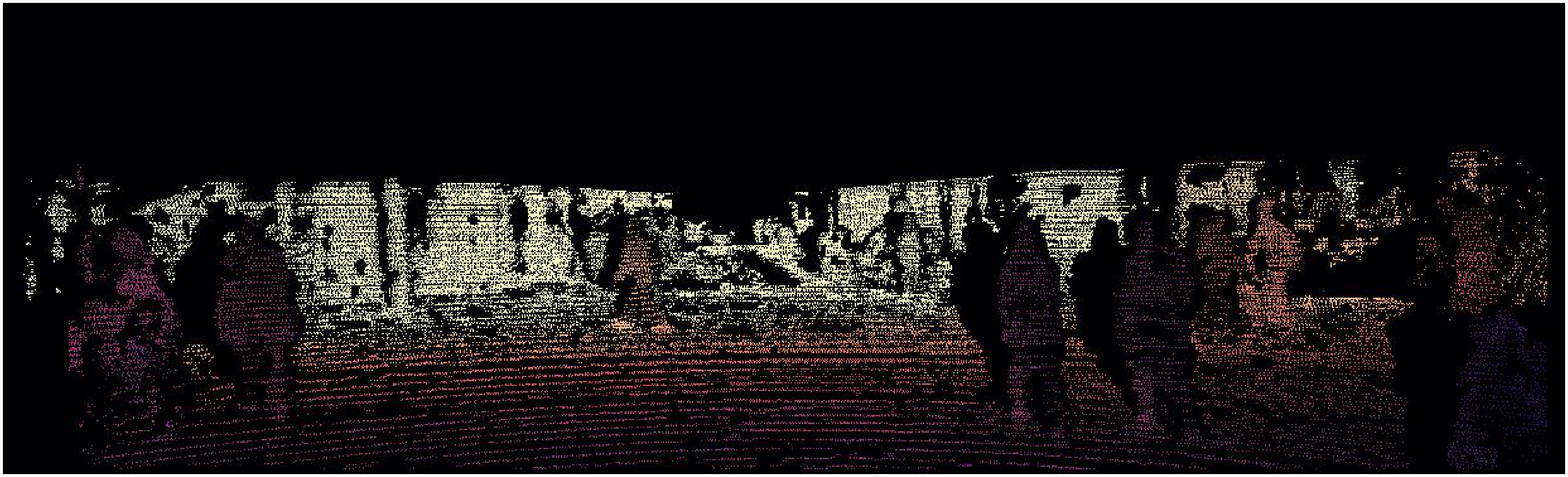}} \\
\multicolumn{1}{c}{\includegraphics[width= 0.24 \linewidth]{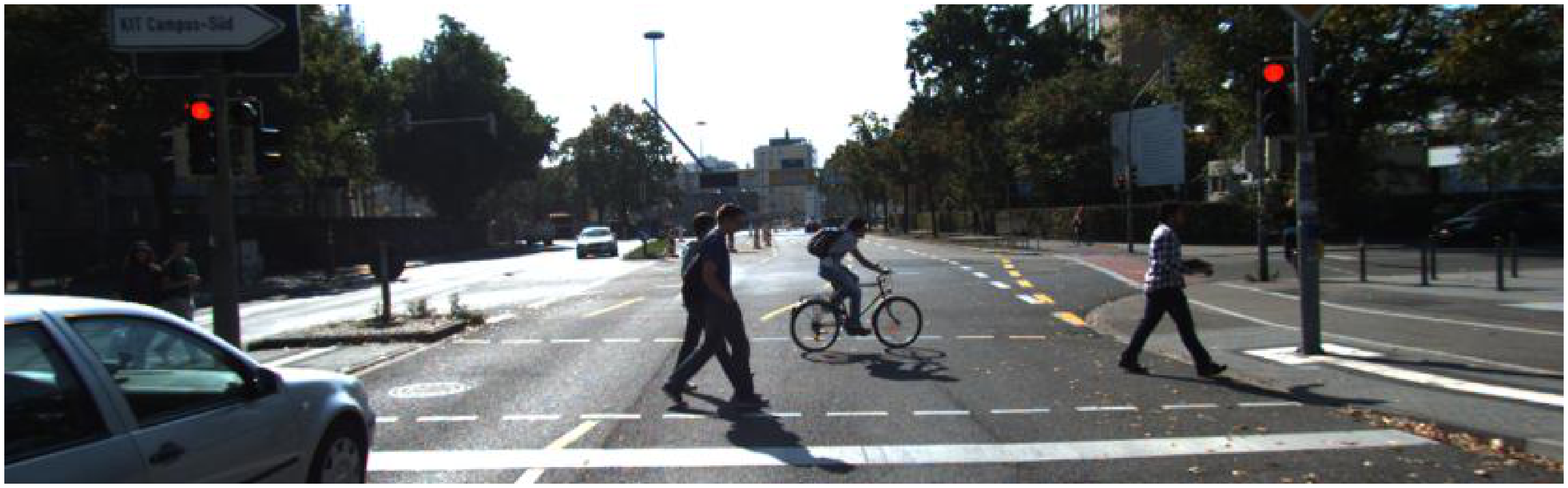}} & \multicolumn{1}{c}{\includegraphics[width= 0.24 \linewidth]{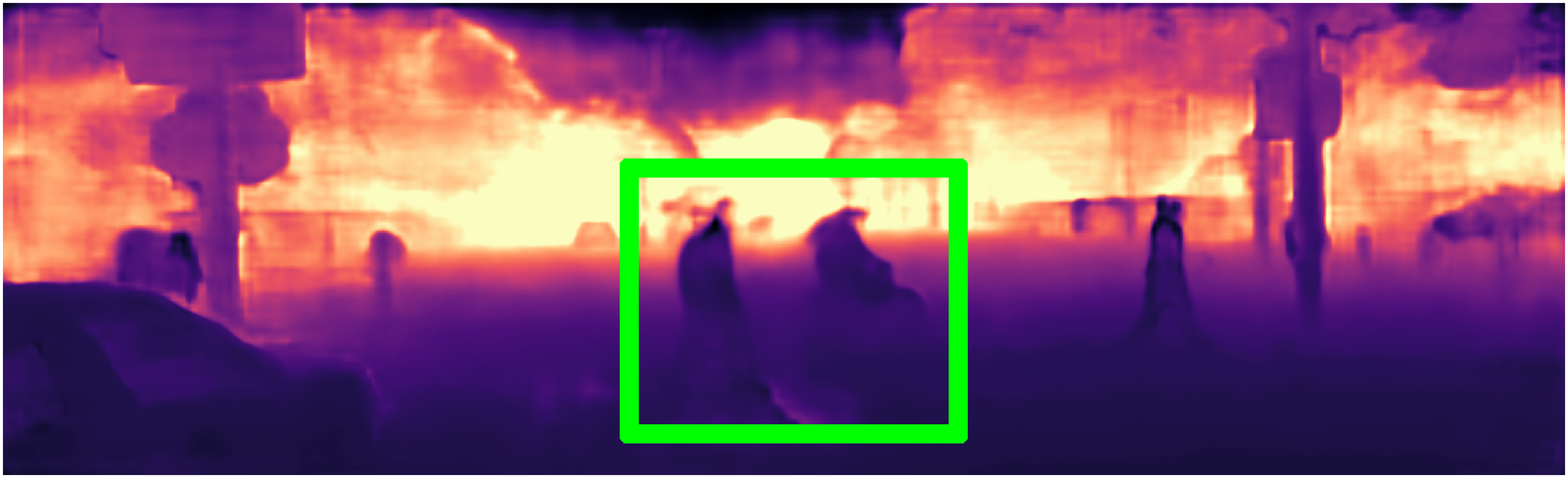}} &
\multicolumn{1}{c}{\includegraphics[width= 0.24 \linewidth]{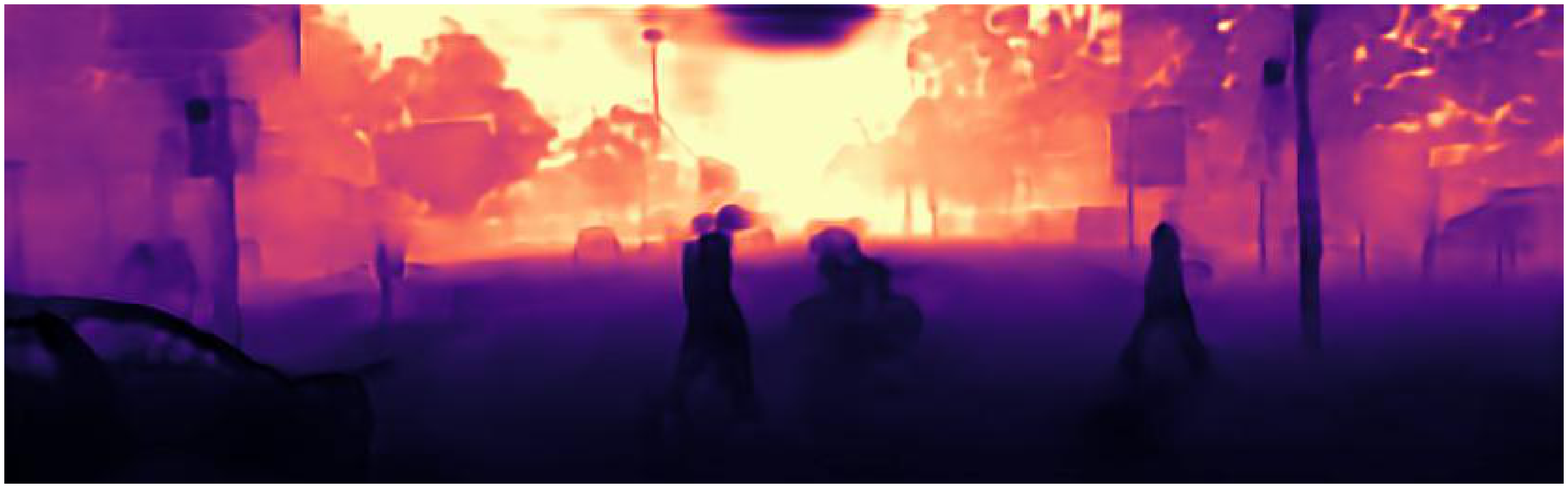}} & \multicolumn{1}{c}{\includegraphics[width= 0.24 \linewidth]{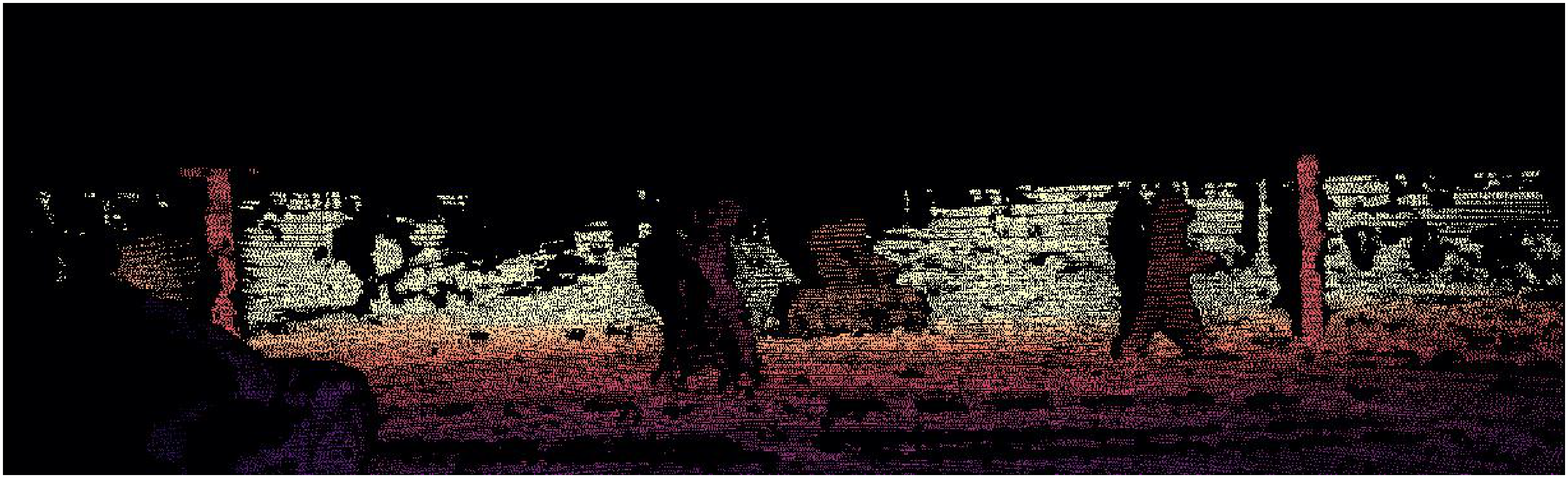}} \\
\multicolumn{1}{c}{\includegraphics[width= 0.24 \linewidth]{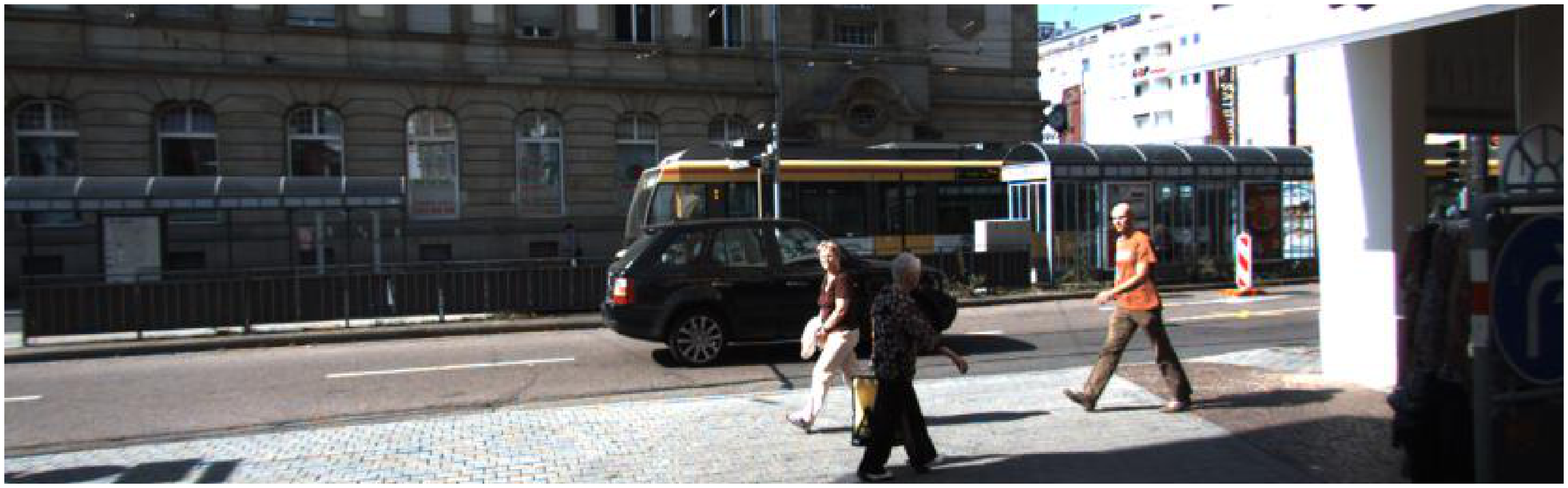}} & \multicolumn{1}{c}{\includegraphics[width= 0.24 \linewidth]{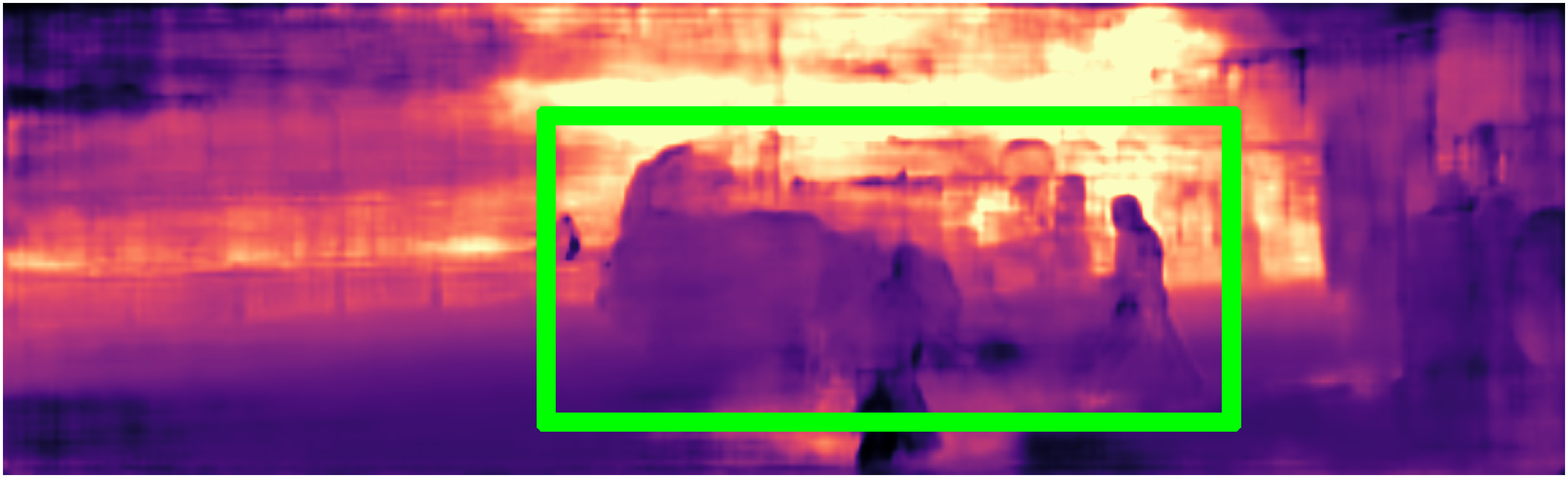}} &
\multicolumn{1}{c}{\includegraphics[width= 0.24 \linewidth]{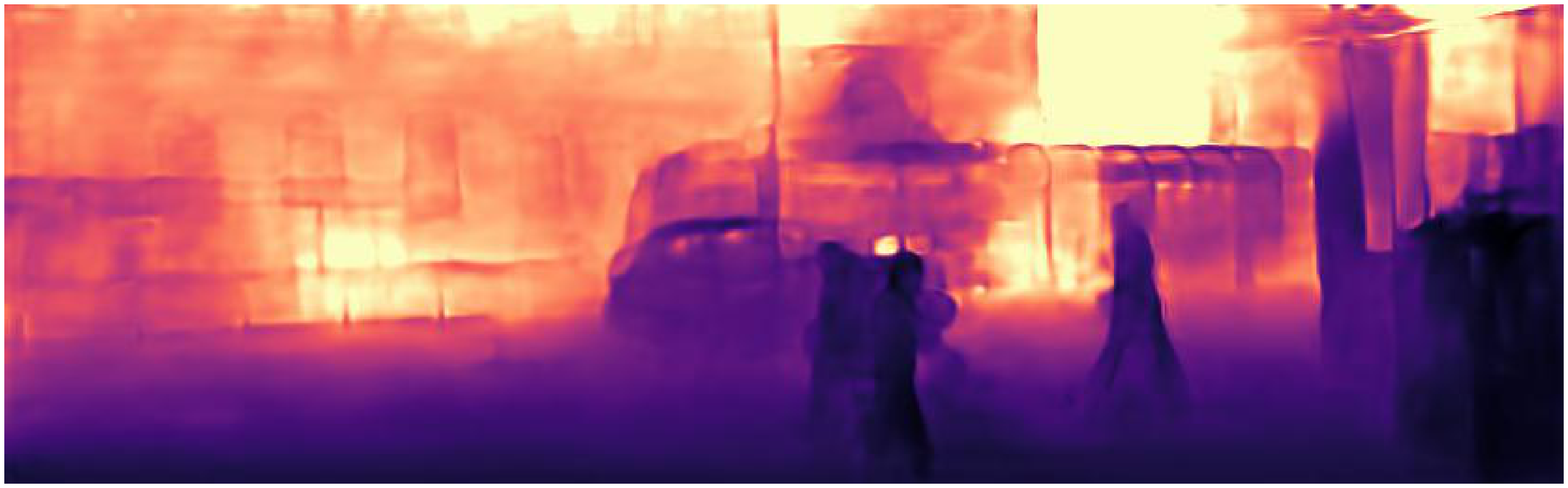}} & \multicolumn{1}{c}{\includegraphics[width= 0.24 \linewidth]{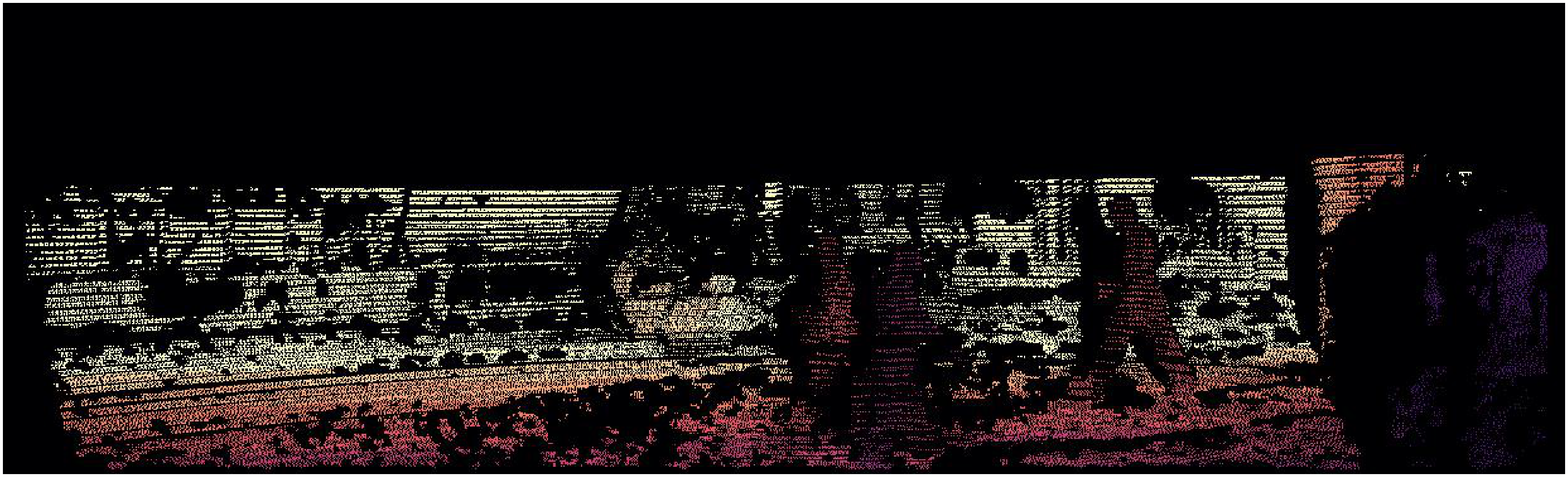}} \\
\multicolumn{1}{c}{\includegraphics[width= 0.24 \linewidth]{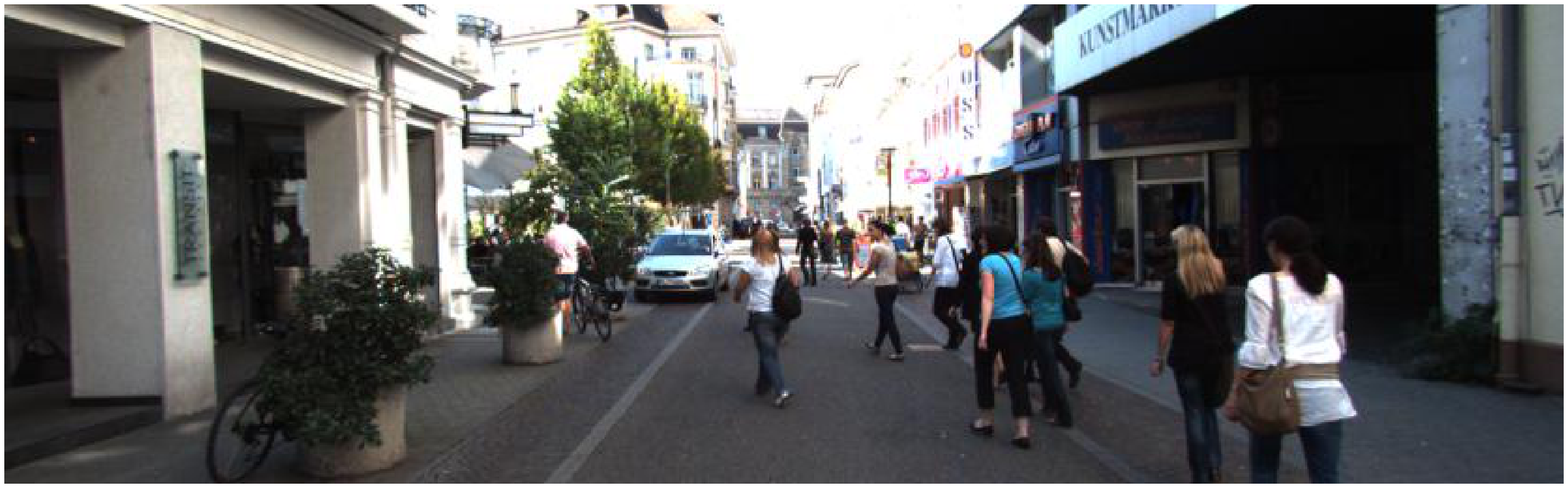}} & \multicolumn{1}{c}{\includegraphics[width= 0.24 \linewidth]{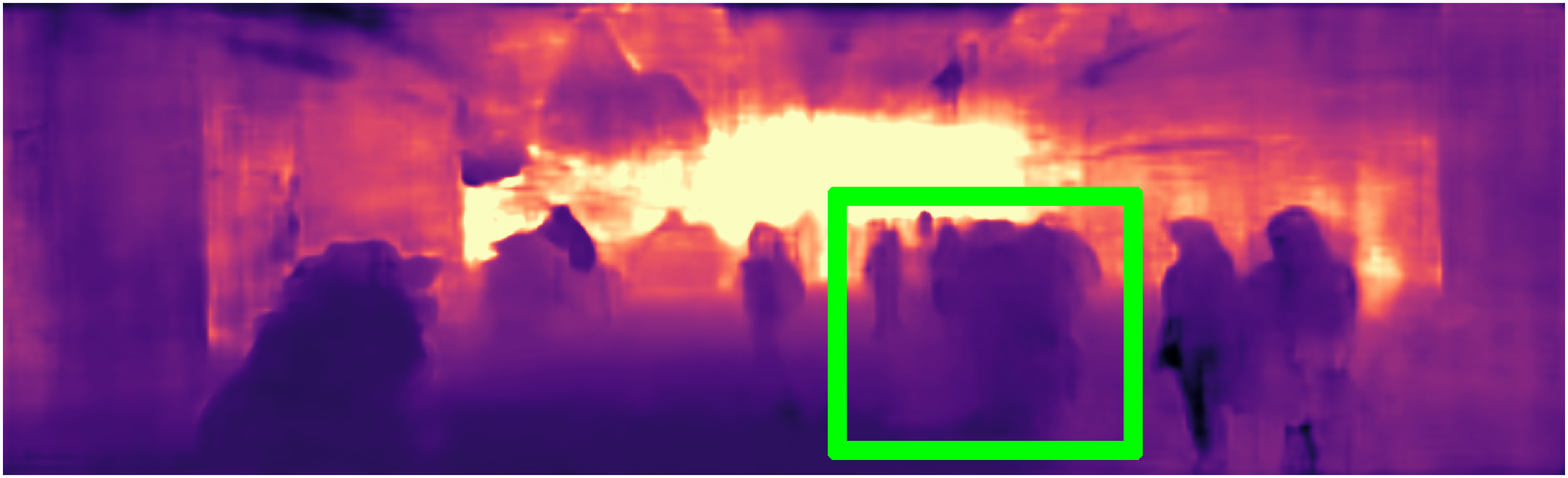}} &
\multicolumn{1}{c}{\includegraphics[width= 0.24 \linewidth]{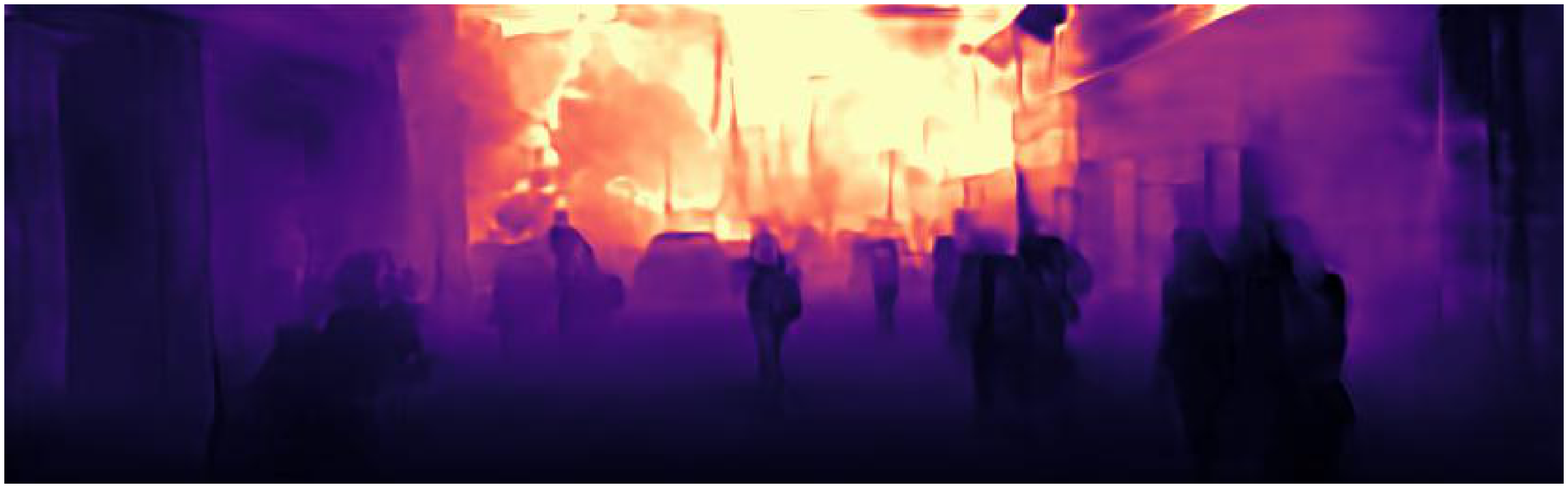}} & \multicolumn{1}{c}{\includegraphics[width= 0.24 \linewidth]{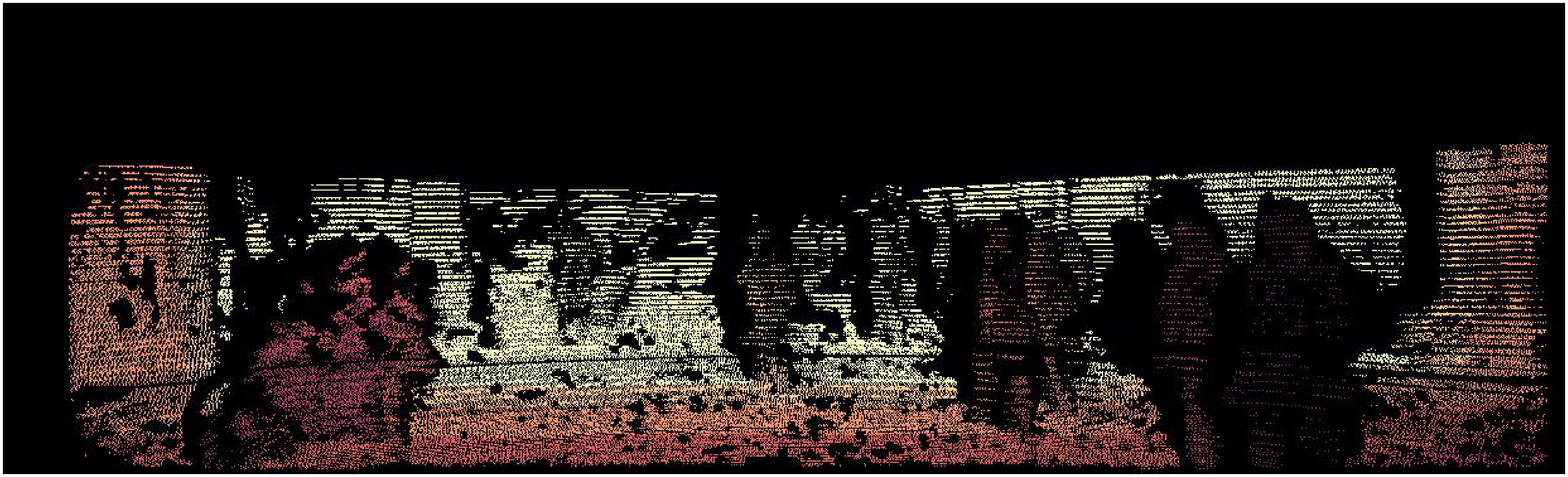}} \vspace{-0.25em}\\
(a) Target image & (b) Wang \emph{et al.} & (c) DeepMLE & (d) Ground truth
\vspace{-0.25em}
\end{tabular} 
\caption{The depth maps generated by the proposed DeepMLE and Wang \emph{et al.} on KITTI dataset. These four scenes above include a lot of dynamic objects. The green boxes are applied to highlight the areas where depth are more accurately estimated by our DeepMLE.}
\vspace{-0.5em}
\label{Fig:results}
\end{figure*}
%

%------------------------------------------------------------------
\vspace{-0.6em}
\subsection{Accuracy Evaluation}
%------------------------------------------------------------------
% evaluate the its estimations of depth and relative pose qualitatively and quantatively.
% This experiment consists of two parts: 1) the depth evaluation on Eigen split\cite{eigen_split}. 2) the pose evaluation on KITTI odometry.

$\textbf{KITTI.}$ To evaluate the accuracy of the estimations of our method, our DeepMLE was tested on the KITTI for the separate evaluation of the estimation of the depth and the relative camera pose.  

For the depth evaluation, we applied the metrics presented in~\cite{eigen2014depth} for fair comparison. Firstly, the qualitative results are presented in Fig.~\ref{Fig:results}. The qualitative results clearly show that compared with the state-of-the-art approach, our DeepMLE performs more robustly in the scenes full of the dynamic objects, and these results owe to the proposed MLE inference. Secondly, the quantitative results are reported in Table~\ref{tab:kitti_depth}. Our DeepMLE has achieved a marked increase for two-view SfM framework. Compared to DeepV2D, which uses a sequence of frames including two closest frames before and after the target frame for training and inferring, our method only takes two consecutive image as input. But our method outperforms DeepV2D by a clear margin. The marked accuracy improvement proves the significance of the effective integration of MLE inference to our model. Meanwhile, compared to Wang ~$\emph{et al.}$~\cite{Wang_2021_CVPR}, the state-of-the-art $\textbf{Type I}$ algorithm, our DeepMLE has achieved competitive results and have outperformed it in four metrics clearly. And in \cite{Wang_2021_CVPR}, the depth is estimated by an isolated module which is supervised by depth ground truth directly, which is more easy to fit ground truth in one data.
\renewcommand{\multirowsetup}{\centering} 
\begin{table*}[!tb]
\small
\tabcolsep=0.125cm
\centering
\caption{\small{Depth evaluation on KITTI dataset.
The \textbf{S} and \textbf{SS} are short for ``Supervised'' and ``Self-Supervised'' respectively.}} 
\begin{tabular}{c| c | c || c | c | c | c | c | c | c }
\hline
\multirow{2}{*}{Split}& \multirow{2}{*}{Types} & \multirow{2}{*}{Methods}  &  \multicolumn{4}{c|} { Lower is better } & \multicolumn{3}{ c  }{ Higher is better }	  \\ 
\cline{4-10}
& & & Abs Rel & Sq Rel & $\text{RMSE}$ & $\text{RMSE}_{log}$  & $\delta < 1.25$ & $\delta < 1.25^{2}$ & $\delta < 1.25^{3}$   \\
\hline\hline
\multirow{9}{*}{\rotatebox{90}{{Eigen}}} & \multirow{3}{*}{\textbf{Type I}}
& BA-Net \cite{BAnet} & 0.083& - & 3.640 & 0.134 & - & - & - \\
\cline{3-10}
&& DeepV2D \cite{DeepV2D} & 0.064 & 0.350 & 2.946 & 0.120 & 0.946 & 0.982 & 0.991\\
\cline{3-10}
&& Wang et al.\cite{Wang_2021_CVPR} & \textbf{0.055} & \textbf{0.224} & \textbf{2.273} & \textbf{0.091} & \textbf{0.956} & \textbf{0.984} & \textbf{0.993}\\
\cline{2-10}

&\multirow{4}{2.2cm}{\textbf{Type II (SS)}} & SfMLearner~\cite{geoconstrains} &0.208 &1.768 &6.856 &0.283 & 0.678 &0.885 &0.957\\
\cline{3-10}
& &monodepth2~\cite{monodepth} & 0.115 & 0.903 & 4.863 & 0.193 & 0.877 & 0.959 & 0.981\\
\cline{3-10}
& &GLNet~\cite{GLnet} &0.099 &0.796 &4.743 &0.186&0.884 &0.955 &0.979\\
\cline{3-10}
& &D3VO\cite{D3vo} & - & - & 4.485 & 0.185 & 0.885 & 0.958 & 0.979\\
\cline{2-10}
&\multirow{3}{2.2cm}{\textbf{Type II (S)}} & DeepMLE (Ours w/o uncertainty) & 0.065 &0.212 &2.628&0.095&0.955&0.984&0.994\\
\cline{3-10}
&& DeepMLE (Ours w/o MLE) & 0.091 &0.367 &3.257&0.129&0.920&0.985&0.996\\
\cline{3-10}
& & DeepMLE (Ours full)& \textbf{0.060} & \textbf{0.203} & \textbf{0.257} & \textbf{0.089} & \textbf{0.967} & \textbf{0.995} &\textbf{0.999}\\
\hline
\end{tabular}
\vspace{-1.0em}
\label{tab:kitti_depth}
\end{table*}

\begin{figure}[t]
\centering
\renewcommand{\tabcolsep}{2pt}
\begin{tabular}{cc}
\multicolumn{1}{c}{\includegraphics[width= 0.50 \linewidth]{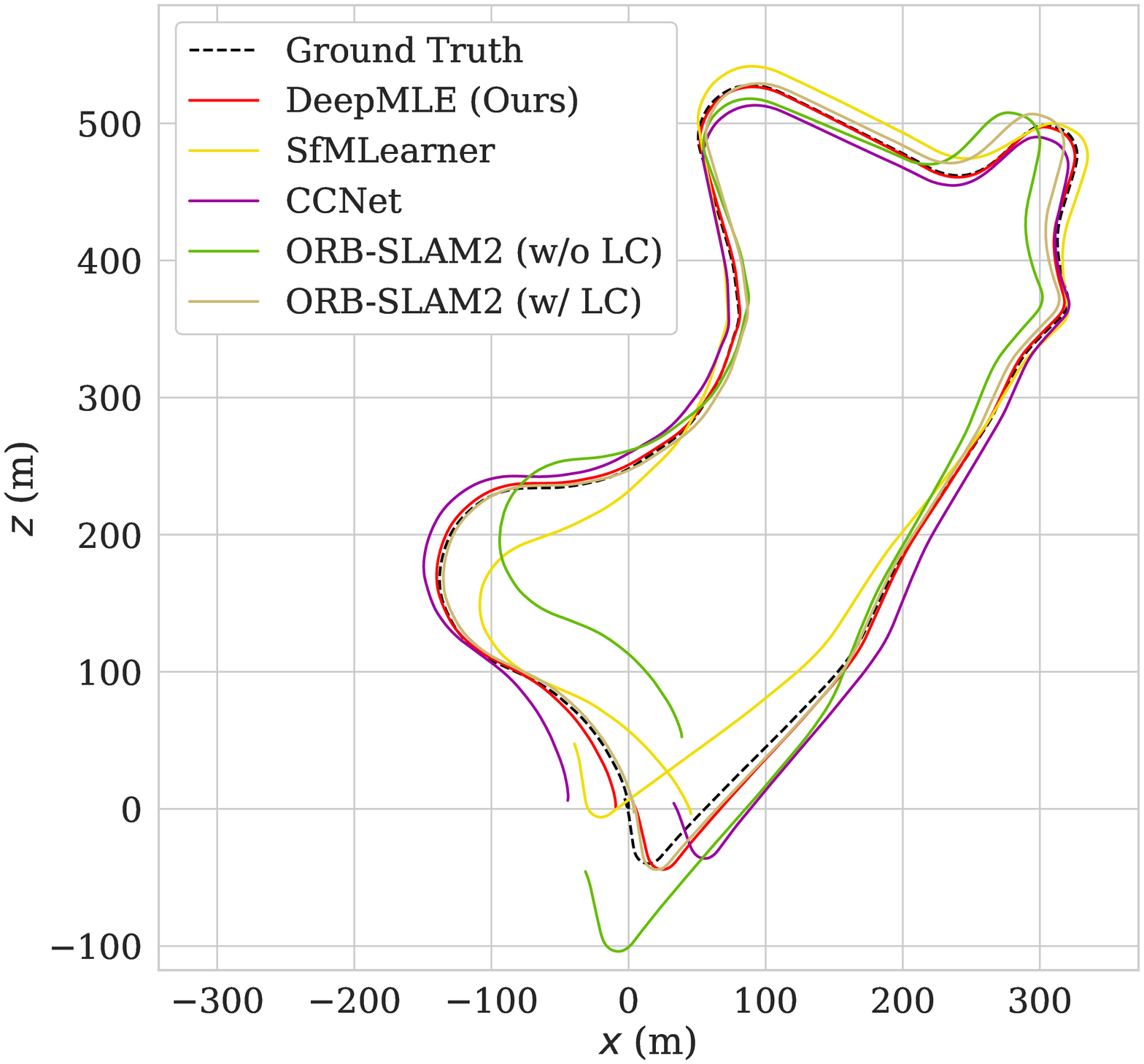}} & 
\multicolumn{1}{c}{\includegraphics[width= 0.50 \linewidth]{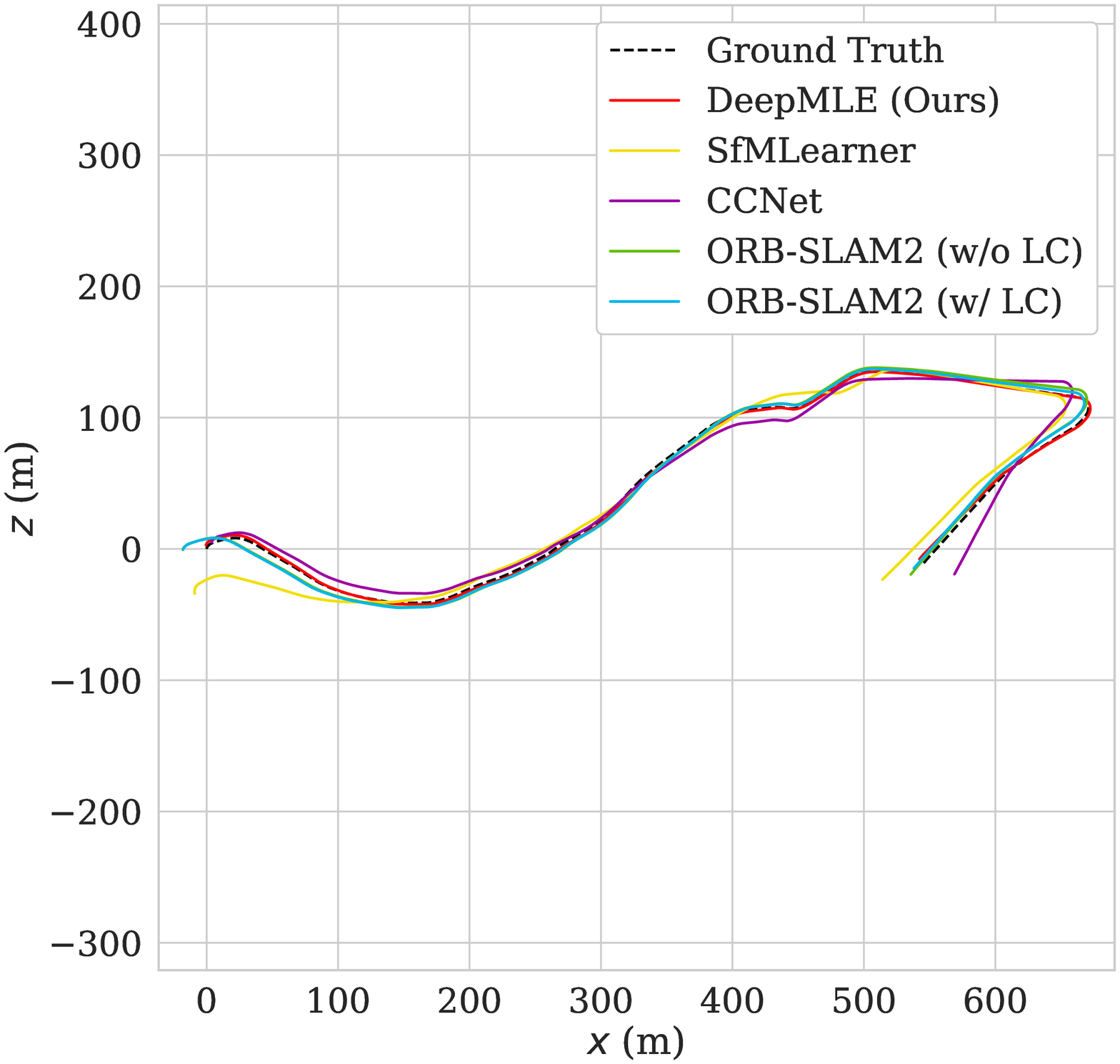}}
\end{tabular} 
\caption{The visualization of trajectory. Left: the visualization in sequence ``09''; Right: the visualization in sequence ``10''}
\vspace{-2.12em}
\label{Fig:traj}
\end{figure}

As the quantitative results shown in Table~\ref{tab:kitti_pose}, our method make a obvious progress for $\textbf{Type II}$ two-view SfM method. Actually, learning to estimate relative camera pose directly by a neural network is much more challenging than estimating it by robust geometry algorithms. But our pipeline increases the accuracy of $\textbf{Type II}$ methods by an order of magnitude. And our method achieves the competitive results with the ORB-SLAM2 with loop closure and back-end optimization. We visualized the trajectories of ``09'' and ``10'' sequences generated by different methods in Fig.\ref{Fig:traj}.

\begin{table}[t]
\begin{center}
\tabcolsep=0.06cm
\caption{\small{Pose estimation accuracy on the KITTI VO dataset.}}
\label{tab:kitti_pose}
\footnotesize
\begin{tabular}{c|c|c |c|c}
\hline
 \multirow{2}{*}{Methods}  & \multicolumn{2}{c|}{Seq. 09} & \multicolumn{2}{c}{Seq. 10} \\
\cline{2-5}
&$t_{\text{err}}  (\%)$ &$r_{\text{err}} (^\circ/100m)$ &$t_{\text{err}}  (\%)$ &$r_{\text{err}} (^\circ/100m)$ \\
\hline
ORB-SLAM2 (wo/ LC)~\cite{orbslam2} &9.30&0.26&2.57&0.32 \\
ORB-SLAM2 (w/ LC)~\cite{orbslam2} &2.88&\textbf{0.25}&3.30&\textbf{0.30} \\
SfMLearner~\cite{geoconstrains}& 8.28 &3.07&12.20 &2.96\\
DeepV2D~\cite{DeepV2D}&8.71&3.70&12.81&8.30 \\
LTMVO~\cite{LTMVO}&3.49&1.03&5.81&1.81\\
Zhao $\emph{et al.}$~\cite{withoutposenet}&6.81&0.72&4.39&1.05\\
TartanVO~\cite{tartanvo2020corl}&6.00&3.11&6.89&2.73
\\
CCNet~\cite{CCNet}&4.32 &1.69& 5.35 &3.16 \\
DeepMLE (Ours)& \textbf{1.46} & 0.76 & \textbf{1.28} & 0.67 \\
\hline
\end{tabular}
\vspace{-1em}\end{center}
\end{table}

$\textbf{DeMoN dataset.}$ We further conducted the experiments on DeMoN dataset, which contains comprehensive scenes of outdoor and indoor. For comparison, we directly borrow the results from the corresponding articles~\cite{DeepSFM,Wang_2021_CVPR}. Both traditional~\cite{COLMAP} and learning-based methods~\cite{BAnet,DeepSFM,DeepV2D,Wang_2021_CVPR,LsNet} are selected for comparison. In this dataset, we tested all methods on SUN3D, RGB-D and Scenes11 respectively. The evaluation results are presented in Table~\ref{table:DeMoN}. Although DeepSFM utilized the initialized depth map and relative camera poses estimated by DeMoN network, Our DeepMLE does not need any initialization and beats it by a clear margin in RGB-D and SUN3D datasets and achieves the similar performance in the Scenes11 dataset.

\begin{table*}[!htb]
\begin{centering}
\tabcolsep=0.08cm
\small
\vspace{0.5em}
\caption{Results on DeMoN datasets. The traditional methods are reported in the first three rows. DeMoN, LS-Net and DeepSFM are the $\textbf{Type II}$  methods. While Wang~\emph{et al.}~\cite{Wang_2021_CVPR} and BA-Net~\cite{BAnet} are $\textbf{Type I}$  methods.}
\label{table:DeMoN}
{
\begin{tabular}{c||ccc|cc||ccc|cc||ccc|cc}
\hline
\multirow{3}{*}{Methods} & \multicolumn{5}{c||}{{\textbf{RGB-D Dataset} }} &\multicolumn{5}{c||}{{\textbf{Scenes11 Dataset}}} &\multicolumn{5}{c}{{\textbf{Sun3D Dataset}}} \\
 \cline{2-16}
% \cline{3-10}
&\multicolumn{3}{c|}{{Depth}} & \multicolumn{2}{c||}{{Pose}}&\multicolumn{3}{c|}{{Depth}} & \multicolumn{2}{c||}{{Pose}}&\multicolumn{3}{c|}{{Depth}} & \multicolumn{2}{c}{{Pose}} \\ \cline{2-16}
& L1-inv &Sc-inv &L1-rel & Rot & Tran &L1-inv &Sc-inv &L1-rel & Rot & Tran&L1-inv &Sc-inv &L1-rel & Rot & Tran\\
\hline
Base-SIFT &0.050 &0.577 &0.703 &12.010& 56.021& 0.051& 0.900& 1.027 &6.179& 56.650 & 0.029& 0.290& 0.286& 7.702& 41.825\\
Base-Matlab & - &- &- &12.813& 49.612&  - &- &-& 0.917 &14.639& - &- &- &5.920& 32.298\\
COLMAP~\cite{COLMAP}&-&-&0.384& 7.961&23.469&-&-&0.625&4.834&10.682 &-&-&0.623&4.235&15.956  \\
\hline
BA-Net~\cite{BAnet} & 0.008 & 0.087 & \textbf{0.050} & 2.459 & 14.900 & 0.080 &0.210& 0.130 &3.499 &10.370 & 0.015 & 0.110 & 0.060 & 1.729 & 13.260 \\
Wang~\emph{et al.}~\cite{Wang_2021_CVPR} &-&-&-& -&-& \textbf{0.005}&\textbf{0.097} & \textbf{0.058} & \textbf{0.276}&\textbf{2.041}&\textbf{0.010}&\textbf{0.081}&\textbf{0.057}&\textbf{1.391} &\textbf{10.757}\\
\hline
DeMoN~\cite{DeMoN} & 0.028 & 0.0.130 & 0.212 & 2.641 & 20.585 & 0.019 & 0.315 & 0.248 & 0.809 & 8.918 & 0.019 &0.114 &0.172& 1.801 & 18.811\\

% \midrule
LS-Net~\cite{LsNet} & 0.019 & 0.090 & 0.301 & \textbf{1.010} & 22.100 &  0.010& 0.410& 0.210 &4.653 &8.210 & 0.015 & 0.189 & 0.650& 1.521& 14.347  \\
DeepSFM~\cite{DeepSFM} &  0.011 &  0.071 &  0.126 &  1.862 &  14.570  &  0.007  & 0.112 &  0.064  & 0.403 &  5.828 & 0.013 & 0.093 & 0.072 & 1.704 & 13.107 \\
DeepMLE (Ours)&\textbf{0.005}&\textbf{0.065}&\textbf{0.068}&2.067&\textbf{13.893}&\textbf{0.005}&0.126&0.079&\textbf{0.386}&\textbf{3.764}&\textbf{0.007}&\textbf{0.091}&\textbf{0.068}&\textbf{1.261}&\textbf{9.763}\\
\hline
\end{tabular}
}
\vspace{-1.0em}
\end{centering}
\end{table*}
%------------------------------------------------------ 
\vspace{-0.5em}
\subsection{Generalization Evaluation}
\label{sec:generalize}
% & & DORN~\cite{DORN} & {0.072} &  {0.307} & {2.727} & {0.120} & {0.932} & {0.984} & {0.994}  \\
% \cline{3-10}
% & & VNL~\cite{VNL} & 0.072&-&3.258&0.117& 0.938 &0.990&0.998\\
% \cline{2-10}
% & &GeoNet~\cite{geonet} &0.155 &1.296 &5.857 &0.233&0.793 &0.931 &0.973\\
% \cline{3-10}
% & &CCNet~\cite{CCnet} &0.140 &1.070 &5.326 &0.217&0.826 &0.941 &0.975\\
% \cline{3-10}
% & &GLNet~\cite{GLnet} &0.099 &0.796 &4.743 &0.186&0.884 &0.955 &0.979\\
% \cline{3-10}
$\textbf{vKITTI to KITTI.}$ The same metrics are used for depth and pose evaluation in generalization experiment. And we followed the experiment settings of the task of domain-adaptation. The vKITTI dataset was taken as the source dataset, while the KITTI was regarded as the target. We trained our DeepMLE in vKITTI, and tested it in KITTI without any fine-tune. As reported in Table~\ref{table:vkitti}, without any operation for domain adaptation, our DeepMLE outperforms the baseline by a clear margin and beats the state-of-the-art domain adaptation method~\cite{S2R_2021_CVPR} in six metrics. The results show that the introduction of MLE inference to deep neural network make the estimation more accurate.

% And it is noteworthy that specific intermediate results and explanation are provided in ~$\textbf{Supplementary Material}$.  
%
% For comparison, we report the state-of-the-art domain-adaptation method in depth estimation\cite{S2R_2021_CVPR}.
\begin{table*}[h!t]
\small
\vspace{0.5em}
\tabcolsep=0.15cm
\centering
\caption{\small{Generalization evaluation of depth estimation from vKITTI to KITTI. We compared our method with the state-of-the-art domain adaptation method S2R-DepthNet~\cite{S2R_2021_CVPR}. And we take a U-Net-like network as baseline which is trained by direct regression. Under the same hardware condition to us, we tested Wang~\emph{et al.}~\cite{Wang_2021_CVPR} method by their published codes.}}
%\resizebox{\textwidth}{!}
{%
\begin{tabular}{c||c|c|c|c|c|c|c}
\hline
 &   \multicolumn{4}{c|}{Lower is better} & \multicolumn{3}{c}{Higher is better} \\  \cline{2-8}
\multirow{-2}{*}{Settings} & Abs Rel & Squa Rel & RMSE & RMSE$_{log}$ & $\delta < 1.25$ & $\delta < 1.25^2$ & $\delta < 1.25^3$ \\ 
\hline
T$^2$Net\cite{T2Net} (vKITTI$\to$KITTI)  & 0.171 & 1.351 & 5.944  & 0.247 & 0.757  & 0.918   & 0.969  \\
S2R-DepthNet\cite{S2R_2021_CVPR} (vKITTI$\to$KITTI) & 0.165 & 1.351 & 5.695  & 0.236 & \textbf{0.781}  & 0.931   & 0.972 \\
baseline (vKITTI$\to$KITTI) &0.236&2.171&7.063&0.315&0.642&0.861&0.944\\
Kundu \emph{et al.}\cite{kundu}(vKITTI$\to$KITTI)&0.214&1.932&7.157&0.295&0.665&0.882&0.950 \\
Wang~\emph{et al.}\cite{Wang_2021_CVPR} (vKITTI$\to$KITTI) &0.245&2.032&7.953&0.305&0.673&0.871&0.953
\\
\hline
DeepMLE (Ours w/o uncertainty) (vKITTI$\to$KITTI)&0.169&1.070&5.396&0.227&0.748&0.930&0.978\\
DeepMLE (Ours w/o MLE) (vKITTI$\to$KITTI)&0.209&1.342&5.518&0.260&0.679&0.908&0.971\\
DeepMLE (Ours full) (vKITTI$\to$KITTI) & \textbf{0.164} & \textbf{1.005} & \textbf{5.135} & \textbf{0.218} & 0.773 & \textbf{0.939} & \textbf{0.980}   \\
\hline
\end{tabular}%
\vspace{-1.0em}
\label{table:vkitti}
}
\end{table*}

$\textbf{MVS dataset.}$
To further illustrate the generability ability of DeepMLE, we trained our method in SUN3D, Scenes11 and RGB-D and tested it in MVS dataset. It should be noted that the MVS dataset is very different to our training data. This dataset has been widely used for the generalization evaluation for SfM problem. As results shown in Table~\ref{table:MVS}, our DeepMLE beats DeepSFM~\cite{DeepSFM}, the state-of-the-art $\textbf{Type II}$ SfM approach, by a clear margin, e.g., 0.014 versus 0.021 in L1-inv and 0.103 versus 0.129 in Sc-inv.

\begin{table}[t]
\centering\small
\begin{center}
\caption{Generalization evaluation on MVS dataset.}
\label{table:MVS}
\tabcolsep=0.1cm
{
\begin{tabular}{c||ccc|cc}
\hline
\multirow{3}{*}{Method} & \multicolumn{5}{c}{{\textbf{MVS dataset} }}  \\
\cline{2-6}
&\multicolumn{3}{c|}{{Depth}}&\multicolumn{2}{c}{{Pose}}\\
&L1-inv &Sc-inv &L1-rel &Rot &Tran\\
\hline
Base-SIFT & 0.056 & 0.309 & 0.361 &21.180 & 60.516
\\
Base-Matlab & -& -& -&10.843 &32.736\\
COLMAP~\cite{COLMAP}&-&-&0.384&7.961&23.469\\
\hline
BA-Net~\cite{BAnet}&0.030&0.150&0.080&3.499&11.238\\
Wang~\emph{et al.}~\cite{Wang_2021_CVPR}&\textbf{0.015}&\textbf{0.102}&\textbf{0.068}&\textbf{2.417}&\textbf{3.878} \\
\hline
DeMoN~\cite{DeMoN}&0.047&0.202&0.305&5.156&14.447\\
LS-Net~\cite{LsNet}&0.051&0.221&0.311&4.653&11.221\\
DeepSFM~\cite{DeepSFM}&0.021&0.129&0.079&2.824&9.881\\
DeepMLE (Ours)&\textbf{0.014}&\textbf{0.103}&\textbf{0.072}&\textbf{1.261}&\textbf{9.762}\\
\hline
\end{tabular}
\vspace{-2.5em}
}
\end{center}
\end{table}

%------------------------------------------------------------------------
\vspace{-0.5em}
\subsection{Ablation Study}
%------------------------------------------------------------------------

We conducted a set of experiments to justify the designations of each part of DeepMLE. The results of ablation studies are presented in the last three rows of Tables~\ref{tab:kitti_depth} and~\ref{table:vkitti}.

$\textbf{Uncertainty module.}$ Taking no consideration of the uncertainty, we use the fixed parameters for the Gaussian and Uniform mixture distribution instead using the parameters predicted by proposed uncertainty module. The results presented in Tables~\ref{tab:kitti_depth} and~\ref{table:vkitti} shows that without the uncertainty module, the accuracy and generability capability of our network distinctly drop. These result further underlines the importance of the predictive uncertain parameters which alleviate the influence of illumination changes, occlusions and image noise.

%help the network learn the better correlation which more conforms to physical situation and increase the robustness of our algorithm.

%The result in the last two rows of Table \ref{tab:kitti_depth} shows the comparison of DeepMLE with uncertainty prediction and without uncertainty.

$\textbf{MLE inference.}$ In order to justify the significance of integrating the MLE inference into DNNs. We trained our framework by ablating the MLE inference parts, and the depth and relative camera pose are regressed from one block instead of iterative inference. And the results shown in Tables~\ref{tab:kitti_depth} and~\ref{table:vkitti}, when the MLE inference is removed, the accuracy and generalization capability sharply drop in all the metrics. These two ablations experiments distinctly justify that the significance of the MLE inference for generalization capability and accuracy of our model.     
% And in table.\ref{table:vkitti}, the result shows that the impact of MLE inference both in the increase of accuracy and generalization.

%------------------------------------------------------------------------
%------------------------------------------------------------------------
\vspace{-0.8em}
\section{CONCLUSIONS}
In this paper, we creatively integrated the MLE inference into the designed deep neural network, which made the end-to-end model more interpretable and genearalizable. Instead of using tradition methods to maximize the likelihood of the observations, we propose an end-to-end network to solve it using gradient-like information. Thus, we can take advantages of both learning-based method and traditional MLE method. Extensive experiments on several datasets illustrates that the proposed DeepMLE has achieved state-of-the-art performance in both generalization capability and accuracy.

%-------------------------------------------------------------------------
%-------------------------------------------------------------------------
\vspace{-0.8em}
\section{ACKNOWLEDGE}
This work was partially supported by the National Natural Science Foundation of China (No. 42101440), the CCF-Baidu Open Fund (No.OF2021023), the Shenzhen Central Guiding the Local Science and Technology Development Program (No.2021Szvup100), and the LIESMARS Special Research Funding.

%%%%%%%%%%%%%%%%%%%%%%%%%%%%%%%%%%%%%%%%%%%%%%%%%%%%%%%%%%%%%%%%%%%%%%%%%%%%%%%%

%%%%%%%%%%%%%%%%%%%%%%%%%%%%%%%%%%%%%%%%%%%%%%%%%%%%%%%%%%%%%%%%%%%%%%%%%%%%%%%%



% \section{ACKNOWLEDGMENT}

%%%%%%%%%%%%%%%%%%%%%%%%%%%%%%%%%%%%%%%%%%%%%%%%%%%%%%%%%%%%%%%%%%%%%%%%%%%%%%%%
{\small
\bibliographystyle{IEEEtran}
\bibliography{refs/egbib.bib}
}

\end{document}